\documentclass[10pt]{article}

\usepackage{amsmath}
\usepackage{amssymb}

\usepackage{graphicx}

\usepackage{cite}

\usepackage{color} 

\usepackage[labelfont=bf,labelsep=period,justification=raggedright]{caption}

\makeatletter
\renewcommand{\@biblabel}[1]{\quad#1.}
\makeatother

\date{}

\usepackage{courier}
\usepackage{longtable}
\usepackage{multirow}
\newcommand{\insthead}{{\sc ip}}
\newcommand{\readhead}{{\sc read} head}
\newcommand{\writehead}{{\sc write} head}
\newcommand{\flowhead}{{\sc flow} head}
\newcommand{\instset}[1]{{\sc #1}}
\newcommand{\instr}[1]{{\tt #1}}
\newcommand{\instrb}[1]{\texttt{\textbf{#1}}}

\begin{document}

\begin{flushleft}
{\Large
\textbf{Understanding Evolutionary Potential in Virtual CPU Instruction Set Architectures}
}
\\
David M. Bryson$^{\ast}$, 
Charles Ofria
\\
BEACON Center for the Study of Evolution in Action and the Department of Computer Science and Engineering, Michigan State University, East Lansing, MI, USA
\\
$\ast$ E-mail: brysonda@egr.msu.edu
\end{flushleft}

\section*{Abstract}

We investigate fundamental decisions in the design of instruction set architectures for linear genetic programs that are used as both model systems in evolutionary biology and underlying solution representations in evolutionary computation. We subjected digital organisms with each tested architecture to seven different computational environments designed to present a range of evolutionary challenges. Our goal was to engineer a general purpose architecture that would be effective under a broad range of evolutionary conditions. We evaluated six different types of architectural features for the virtual CPUs: (1) genetic flexibility: we allowed digital organisms to more precisely modify the function of genetic instructions, (2) memory: we provided an increased number of registers in the virtual CPUs, (3) decoupled sensors and actuators: we separated input and output operations to enable greater control over data flow. We also tested a variety of methods to regulate expression: (4) explicit labels that allow programs to dynamically refer to specific genome positions, (5) position-relative search instructions, and (6) multiple new flow control instructions, including conditionals and jumps. Each of these features also adds complication to the instruction set and risks slowing evolution due to epistatic interactions. Two features (multiple argument specification and separated I/O) demonstrated substantial improvements in the majority of test environments, along with versions of each of the remaining architecture modifications that show significant improvements in multiple environments. However, some tested modifications were detrimental, thought most exhibit no systematic effects on evolutionary potential, highlighting the robustness of digital evolution. Combined, these observations enhance our understanding of how instruction architecture impacts evolutionary potential, enabling the creation of architectures that support more rapid evolution of complex solutions to a broad range of challenges.

\section*{Introduction}

Over the past 50 years, the field of evolutionary computation has produced many successful tools, including genetic algorithms \cite{Holland1975}, genetic programming \cite{Koza1990}, and evolutionary strategies \cite{Rechenberg1971} (for a recent overview, see \cite{Simon2013}). These evolutionary algorithms abstract the evolutionary process by alternating between selecting the most promising prospective solutions from a diverse population, and randomly mutating copies of those solutions to create further diversity. Evolutionary algorithms now rival human designers in wide-ranging problem domains, from controlling finless rockets \cite{GomezMiikkulainen2003} to automatically patching software bugs \cite{ForrestNguyenWeimer2009}. However, these methods abstract the evolutionary process and tend to be limited in the complexity of the solutions they produce while also losing some of the inherent robustness that occurs in naturally evolved organisms.

Digital evolution is a type of linear genetic programming that provides a rich environment to study evolution in a more natural environment; populations of self-replicating computer programs must survive in a computational world where they are subject to mutations, environmental effects, interactions with other programs, and the pressures of natural selection~\cite{OfriaBrysonWilke2009}. These ``digital organisms'' evolve in more of an unconstrained manner, enabling biologists to explore questions that are difficult or impossible to study in natural systems (e.g., \cite{ChowWilkeOfria2004,CluneMisevicOfria2008,LenskiOfriaPennock2003,WilkeWangOfria2001}). In turn, these more nuanced systems have proven their ability to come up with effective algorithms for practical applications, such as distributed problem solving~\cite{KnoesterMcKinleyBeckmann2007,BeckmannMcKinleyKnoester2007}, software models for dynamic systems~\cite{GoldsbyKnoesterCheng2007}, and robot movement and decision making~\cite{GrabowskiElsberryOfria2008,GrabowskiBrysonPennock2010,Grabowski:2011ur,Grabowski:2013fu}. In short, digital evolution is becoming an essential model system for studying evolutionary mechanisms, while discerning these natural processes is equally crucial to constructing flexible and resilient computing systems~\cite{McKinleyChengOfria2008}.

The instruction set architecture is the core of every instance of digital evolution, defining the characters and syntax of the genetic language, as well as the virtual hardware upon which that language executes. The design of the instruction set architecture within an evolvable system plays an important role in influencing the robustness and flexibility of evolved solutions~\cite{OfriaAdamiCollier2002}.  As the scope and complexity of research performed using digital evolution expands, it is important to ensure that our language is as general purpose as possible, as well as to understand how changes to architecture impact the evolutionary potential of the system. Our previous work has shown that digital evolution is surprisingly robust to poor design decisions \cite{BrysonOfria2012}. Here we have investigated a series of engineered instruction set architecture modifications built upon the underlying von Neumann architecture of Avida, progressively identifying and integrating architectural features that enhance evolutionary potential. In order to test the effect of each modification, we utilized seven computational environments representing a wide range of desired capabilities for solving primarily static optimization problems. We evaluate the final results of experiments performed in each environment with each instruction set modification.

\section*{Methods}

We performed all experiments using executables based on Avida version 2.12, with modifications to support each of the new instruction set architectures that we investigated\footnote{Avida 2.12 source code is available for download, without cost, from http://avida.devosoft.org/.}.  We tested each instruction set architecture with 200 replicate populations in each of seven computational environments. The populations consisted of 3,600 individuals on a 60x60 toroidal grid, and were run for 100,000 updates, where an update is a unit of time in Avida equal to an average of 30 instructions executed per living organism; in practice this translates to a widely varying number of generations depending on the evolved complexity of the digital organisms (somewhere between 500 and 100,000 generations; a mean of 12,423 for the experiments presented here). Organisms were subject to mutations at a standard substitution rate of $2.5 \times 10^{-3}$ per site in the genome, along with a $5 \times 10^{-4}$ probability each for a single instruction insertion or deletion per site in the genome.  All substitutions, insertions, and deletions occurred upon division of the offspring.  We seeded each population with a single ancestral organism capable only of self-replication. Small variations in the initial genotype used in each architecture were often necessary, due to functional differences among the instruction sets, but we limited these variations specifically to neutral labeling instructions (nop-sequences, as described below) used in self-replication.

All statistical tests were conducted using MATLAB 2012a. Configuration files, analysis scripts, and experimental results are available from figshare\cite{BrysonOfria2013:Data}: http://dx.doi.org/10.6084/m9.figshare.826206

\subsection*{Instruction Set Architectures} 

\begin{figure}[!ht]
\begin{center}
\includegraphics{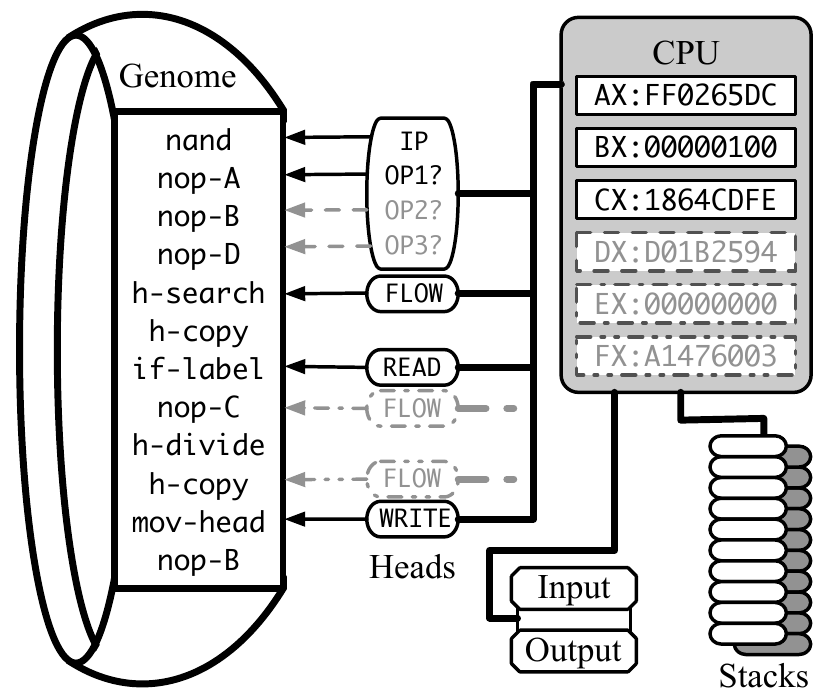}
\end{center}
\caption{\textbf{The architecture of the Avida virtual CPU.} Registers (upper right), stacks (lower right), genomic program (left), heads (middle), and environmental channels (lower right). The solid lines depict the default \instset{Heads} architectural features. The dashed lines show some of the modifications tested.}
\label{fig:architecture}
\end{figure}

The \instset{Heads} instruction set architecture in Avida is the default virtual CPU configuration, consisting of Turing complete, von Neumann style architecture.\footnote{All versions of Avida 2.x through version 2.12 share a functionally equivalent default virtual CPU configuration.  Some specific details in the configuration files have changed across versions.} The virtual hardware that implements this instruction set is designed to operate on a genomic program within a circular memory space (as shown on the left side of Figure \ref{fig:architecture}). By default, it has three registers, each capable of holding a 32-bit number, two stacks that can each hold ten values, four heads that point to positions in the genome, input and output channels, and the ability to execute 26 standard instructions (see Table \ref{tbl:is:insts} for a complete glossary of instructions). The default instructions include three no-operation instructions (nops): \instr{nop-A}, \instr{nop-B}, and \instr{nop-C}, which can serve to modify the default behavior of other instructions, but do not otherwise affect the state of the virtual CPU when executed by themselves. Most instructions observe the value of one subsequent nop instruction and alter their behaviors accordingly. For example, the \instr{inc} instruction increments the BX register by default, but if it were followed by a \instr{nop-A} it would increment the AX register instead. In addition to instruction modification, nop instructions can serve as patterns that act as labels for genome locations. Label matching uses cyclic complementary matching, where \instr{nop-A} matches to \instr{nop-B}, \instr{nop-B} matches to \instr{nop-C}, and \instr{nop-C} matches to \instr{nop-A}.

The \instset{Heads} instruction set has five flow-control instructions: \instr{h-search}, \instr{jmp-head}, \instr{mov-head}, \instr{get\-head}, and \instr{set-flow}. Each of these instructions can affect the position of one of the four architectural heads: the instruction pointer (\insthead), \readhead, \writehead, and \flowhead. The \instr{h-search} instruction searches the genome, starting from the first executed instruction in the genome, for a label (a sequence of one or more nop instructions) that matches the cyclic complementary label that follows the instruction, placing the \flowhead~after the matching sequence; if the sought-after label is not found, it places the \flowhead~on the instruction immediately subsequent to itself. Thus if the \instr{h-search} instruction were followed by \instr{nop-A nop-A nop-B} it would search for the genome for the sequence \instr{nop-B nop-B nop-C}. This is one of only two instructions in the default \instset{Heads} instruction set that is affected by more than one nop instruction, the other being \instr{if-copied} described below. The \instr{mov-head} instruction moves the \insthead~to the current location of the \flowhead. The \instr{jmp-head} instruction shifts the position of the \insthead~by the amount specified in a register. The \instr{get-head} instruction places the current location of the \insthead~into a register. Finally, the \instr{set-flow} instruction moves the \flowhead~to the absolute genome location specified by the value in a register.

The \instset{Heads} set also contains three conditional instructions that will skip a subsequent instruction if the test condition is false. The two basic conditional instructions, \instr{if-n-equ} and \instr{if-less}, perform a comparison between two registers. The \instr{if-copied} instruction interacts with the \readhead, evaluating to true if the last sequence of instructions copied matches the complement of the label that follows the instruction. This instruction is primarily for use in conjunction with the replication instructions described below to identify the portion of the genome most recently copied. 

Seven arithmetic and logic operations are supported in the default \instset{Heads} instruction set: \instr{add}, \instr{sub}, \instr{inc}, \instr{dec}, \instr{nand}, \instr{shift-l}, and \instr{shift-r}. All of these instructions operate on values stored within registers and accept a single nop modifier, which changes the source and destination registers depending on the operation.

Five instructions in \instset{Heads} facilitate data movement and environmental interaction. The \instr{push}, \instr{pop}, and \instr{swap-stk} instructions all operate on the two stacks within the architecture. Only one stack is accessible at a time, with the swap-stk instruction toggling the currently active stack, while \instr{push} and \instr{pop} copy numbers from registers to the top of the active stack and vice-versa. Each of these instructions can be nop-modified to specify which register should be used.  The \instr{swap} instruction exchanges the values of two registers. The \instr{IO} instruction interacts with the environment of the digital organism, outputting the current value in a register and replacing it with a value from the environmentally controlled input buffer. Values output via this instruction are evaluated by the environment, potentially triggering a reward or other action if they match one of the tasks in the environment as explained below.

Lastly, there are three instructions that facilitate self-replication. The \instr{h-alloc} instruction allocates additional memory within which the digital organism can copy its offspring. Copying is performed by repeated execution of the \instr{h-copy} instruction, which duplicates the current instruction found at the \readhead~to the position marked by the \writehead~and advances both heads. Once copying has been completed, the organism must execute the \instr{h-divide} instruction to finalize the replication process, extracting the memory between the \readhead~and the \writehead~as the genome of the offspring.

\subsection*{Tested Architecture Modifications}

In the default \instset{Heads} instruction set, most instructions can have one aspect of their function modified by a single nop instruction that follows in the genome. We aimed to improve the flexibility by which data could be accessed and modified in the virtual CPUs by implementing the \instset{Fully-Associative} (\instset{FA}) instruction set. We extended the nop modification system used by instructions so that most instructions could be modified by more than one nop. The default behavior of all instructions remains the same when not followed by any nop instructions. Instructions that affect only a single register or head retain identical behavior to the \instset{Heads} in the presence of a nop.  However, for arithmetic, logic, and conditional instructions that use multiple registers, the \instset{FA} instruction set will shift all registers to correspond with a signal nop given, as well as read subsequent nops, if present, to further specify those parameters. For example, an \instr{add} instruction, by default will perform $regB = regB + regC$. If it is followed by one \instr{nop-A}, this will alter both the source and destination registers such that it performs $regA = regA + regB$. When followed by \instr{nop-A} \instr{nop-C} \instr{nop-B}, the add instruction in the \instset{Fully-Associative} set will perform $regA = regC + regB$. In this manner, very specific operations may be invoked, while retaining robust default behavior.

The \instset{Register}-series of instruction set architectures build upon the \instset{Fully-Associative} architecture to increase the working register set beyond the three default registers, exposing one or more additional architectural registers, in sets \instset{R4}, \instset{R5}, \instset{R6},  \instset{R7},  \instset{R8},  \instset{R12}, up to a total of 16 in \instset{R16}. The original design choice was made to minimize the number of registers in order to simplify the complexity of using them, but a larger number of registers has not previously been systematically tested. For each additional register, we added a corresponding nop instruction to the instruction set (\instr{nop-D}, \instr{nop-E}, \instr{nop-F}, etc.). None of the default registers used by the instruction set were altered, meaning that these additional registers can be accessed only when the new nop instructions are used to modify an instruction. Since nop modification is also used for head selection, the additional \instr{nop-D} in the \instset{R4} architecture provides direct access to the \flowhead.  In the \instset{R5} through \instset{R16} architectures, extra unassigned heads that may be used as genome place-markers are available for each additional nop instruction.

The \instset{Label}-series of instruction set architectures extends the \instset{R6} architecture (which proved to be the most effective, as described in the results below), explicitly separating genome labels from nop sequences used to modify instruction operands.  The intent of this change was to prevent instruction argumentation as facilitated by the \instset{Fully-Associative} architecture from otherwise conflicting with labeled genome positions, especially those used for self-replication. Instructions that operate on genome labels, \instr{search-seq-comp-s} and \instr{if-copied-seq-comp}, were extended with variants (\instr{search-lbl-comp-s} and \instr{if-copied-lbl-comp}) that recognize sequences of nop instructions only if they begin with the special \instr{label} instruction (see Table \ref{tbl:is:lblsets} for details about the specific instructions included in each set). When executed directly, the \instr{label} instruction performs no operation. The \instset{Label-Direct}-series architectures alter the pattern matching algorithm from the default of cyclic-complementary to direct sequence matching. The \instset{Label-Both} architectures include both pattern matching algorithm instruction variants. In order to increase the power of labeled execution flow, all \instset{Label}-series instruction sets omit the \instr{set-flow} instruction that performs absolute addressing.

The \instset{Split-IO} instruction set architecture alters the \instset{Label-Seq-Direct} architecture, splitting the \instr{IO} instruction into two separate \instr{input} and \instr{output} instructions. Both of the new instructions use the same default register location as the \instr{IO} instruction and can each be modified by one nop.

The \instset{Search}-series of instruction set architectures extend the \instset{Split-IO} architecture with enhanced searching and jumping capabilities.  The \instset{Search\-Directional} set adds two pairs of directional \instr{search-} instructions that scan the genome forward or backward relative to the instruction pointer for a label or sequence match. The \instset{Search-Goto} set, adds a single \instr{goto} instruction that reads the nop sequence that follows the instruction, if present, and will unconditionally jump to the first genome location following the matching label that begins with a \instr{label} instruction. If no matching label is found, execution ignores the goto instruction. The \instset{Search-GotoIf} group adds two conditional \instr{goto} variants, \instr{goto-if-n-equ} and \instr{goto-if-less}, that execute the jump only if the conditional test evaluates to $true$.

The \instset{Flow}-series of instruction set architectures builds upon the flow control features of the \instr{Search\-Directional} architecture, testing multiple combinations of additional flow control instructions (Table \ref{tbl:is:flowsets}). The \instset{If0} group adds four single argument \instr{if} instructions, \instr{if-not-0}, \instr{if-equ-0}, \instr{if-gtr-0}, and \instr{if-less-0}, that conditionally execute the following instruction based on the comparison of the argument with $0$. The \instset{IfX} group adds two if variants, \instr{if-gtr-x} and \instr{if-equ-x}, that conditionally execute the following instruction based on the result of comparing $regB$ with a nop modified number. The default value used by \instr{if-gtr-x} and \instr{if-equ-x} is a $1$. For each nop in the label following a given \instr{if-gtr-x} or \instr{if-equ-x} instruction, the bit is left shifted 1, 2, or 3 times for each \instr{nop-B}, \instr{nop-C}, or \instr{nop-D}, respectively. Whenever a \instr{nop-A} is found in the label sequence, the sign-bit of the value is toggled.  Finally, the \instset{MovHead} group adds two conditional \instr{mov-head} variants, \instr{mov-head-if-n-equ} and \instr{mov-head-if-less}, that operate similarly to the conditional \instr{goto} instructions.

\subsection*{Environments}

We use seven distinct computational environments to evaluate the effectiveness of all tested instruction set architectures.  Each environment focuses on a different aspect of the virtual architecture. Environments contain a set of tasks that carry a metabolic reward associated with their performance. These metabolic rewards increase the computation speed of the digital organism's virtual CPU, making it possible to obtain a competitive advantage relative to other organisms in the population. 

The \textbf{Logic-9} environment consists of metabolic rewards for all possible 1- and 2-input binary logic operations; there are 9 unique operations after removing symmetries and the trivial function `echo'.  The tasks are rewarded multiplicatively, thus virtual CPU speed will increase exponentially as additional tasks are performed.  The logic operations are grouped into five reward levels, ranked by difficulty.  The easiest group will double computational speed, while the highest level increases execution speed by thirty-two times.   Each task is rewarded only once during an organisms' lifetime.  This environmental setup is the default for Avida and has been used in most previous experiments (e.g. \cite{LenskiOfriaCollier1999, LenskiOfriaPennock2003, MisevicOfriaLenski2006}).

The \textbf{Logic-77} environment increases the size and complexity of the Logic-9 environment by adding a reward for all 68 unique three-input binary logic operations.  Performance of each of the 77 operations provides an equal benefit, doubling the execution speed of the organism for the first time the computation is performed.

We designed the \textbf{Match-12} environment to test the organisms' ability to build arbitrary numbers, a task that has been observed to be difficult for organisms to perform and confirmed in the experiments described below.  Rewards are granted additively for outputting any or all of twelve possible numbers.  The numbers are spaced approximately exponentially throughout the 32-bit number space, but have no explicit pattern to them.  Each number is rewarded only once during an organisms lifetime.  Near matches are allowed, but the reward decays via a half-life function based upon the number of bits that are incorrect with a minimum threshold of 22 bits correct.

The \textbf{Fibonacci-32} environment rewards organisms multiplicatively for each number in the Fibonacci sequence until the 32nd iteration of the sequence.  After this target, an organism is penalized for additional numbers output, whereby outputting 64 additional numbers will effectively negate all benefit of the first 32.   The purpose of this setup is to examine the capacity of an instruction set to support bounded recursion and conditional looping.

The \textbf{Sort-10} environment supplies a list of 10 random inputs and offers a reward for outputting those values in descending order. Similar to the Match environment, the reward value decays via a half-life function for each incorrectly sorted value, based on the number of moves required to shift it to the correct order.  Given the limited number of available registers in most of the instruction sets we tested, this task requires the use of the stacks and non-trivial flow control.

The \textbf{Limited-9} environment is based on the Logic-9 environment, offering the similar metabolic rewards for all possible 1- and 2-input binary logic operations. However, unlike the Logic-9 environment, a separate, consumable resource is associated with each task. Each of the resources flows into the environment at a fixed rate (100 units per update) and out proportional to current concentration (1\% per update), creating an equilibrium concentration of 10000 units when not consumed by organisms. Organisms may only consume 0.25\% of an available resource at a given time, impacting the actual metabolic reward collected for performing the task associated with that resource. This property of Limited-9 makes it unique among our tested environments. Unlike our other test environments, which represent instances of static optimization,  the fitness landscape of the Limited-9 environment is dynamic. The fitness measurements of a given genotype will be highly dependent upon current resource conditions, and indirect interaction between competing organism niches may lead to ecological complexity.

Finally, the \textbf{Navigation} environment rewards organisms for successfully navigating a circuitous path marked by sign posts, as described in \cite{GrabowskiBrysonPennock2010}. This task requires an organism to use sensors to retrieve a cue from their local grid position and react to that cue by turning left, turning right, moving straight ahead, or repeating the action indicated by the previous cue (requiring the organisms to also evolve memory). Importantly, this environment also tests the robustness of instruction set architectures to the addition of several, experiment specific instructions, in this instance for sensing and moving in the virtual maze. The virtual maze is completely separate from the organism replication space, and varies randomly across replication cycles.

\subsection*{Assessment of Evolutionary Potential} 

We have focused on two measures to evaluate how well populations solved the computational challenges of the environment when evolved with each instruction set architecture: mean fitness and task success.  Both measure ability of the evolved organisms to perform tasks within the environment.

\textbf{Mean fitness} averages the fitness values of each living organism in the population at the moment the experiment finished.  It takes into account both the computational capability of the organism and the efficiency of self-replication.  We examined the distributions of these fitness values for all instruction set variants in each environment. For each modified instruction set, we compared the 200 population fitness values with those of a reference instruction set architecture using a Wilcoxon rank-sum test. We determined significance using $a = 0.05$ with sequential Bonferroni correction. Confidence intervals, as shown in tables below, represent 2.5\% and 97.5\% quantiles that we generated using non-parametric bootstrap with 10,000 iterations. Since all seven environments present metabolic rewards that are exponential (base-2), all fitness values are shown in log$_2$.

\textbf{Task success}, in contrast to fitness, is a direct examination of the computational capabilities of the organisms within the final population, for the specific environment of the experiment. We measure the task success of a population as the sum of the qualities by which the average organism performs each task. To calculate a task success $t_{p}$ of population $p$, we determine each organism's quality at each task and then sum over these values, finally dividing by the total number of organisms in the population.  More formally,
\begin{equation}
t_{p} = \sum_{i=1}^{N_{p}} \sum_{j=1}^{T}\frac{q_{i,j}}{N_{p}}
\label{equ:is:ts}
\end{equation}
where $N_{p}$ is the number of organisms in population $p$, $T$ is the number of tasks in the environment, and $q_{i,j}$ is the quality $q$ at which organism $i$ is performing task $j$. Task quality ($q$) is a value between 0 and 1, where 1 means the organism has found a perfect solution for a task. Environments that support near-matches use task quality to adjust the metabolic reward accordingly. The maximum task success for a given environment is equal to the total number of tasks rewarded in that environment; for example the maximum task success of the Logic-9 environment is nine. Normalized task success, as presented in the following results, divides the observed task success by the maximum in each environment, thus constraining these values to be between zero and one.  Similar to population mean fitness, we compared the distribution of task success of each instruction set to the control architecture using a Wilcoxon rank-sum test, sequential Bonferroni correction, and non-parametric bootstrap confidence intervals.

In most environments task success will be highly correlated with fitness. Since organisms in digital evolution must self-replicate, it is possible for genotypes with identical task success to exhibit vastly different fitness measurements, so both metrics can be informative. Additionally, in some environments task success provides a more consistent measure of the evolutionary potential of the instruction set. For example, in the Limited-9 environment the reduction in resources due to additional task performance may actually reduce average fitness, even though more tasks are being performed.

\section*{Results}

We evaluated each of the six tested types of hardware modifications in consecutive evolutions of the instruction set architecture. The first hardware modification tested was the \instset{Fully-Associative} set, followed by the \instset{Register} sets, \instset{Label} sets, \instset{Split-IO} set, \instset{Search} sets, and finally the \instset{Flow} sets.

\subsection*{Fully-Associative Argumentation}
In conducting our analysis, the \instset{Fully-Associative} (\instset{FA}) instruction set, which addresses the flexibility of register data flow, shows significant improvement in six of the seven environments (Tables \ref{tbl:is:fa:fit} and \ref{tbl:is:fa:ts}). The logic-based environments (Logic-9, Logic-77, and Limited-9) all show substantially improved fitness and task success. The Logic-77 environment in particular, benefits from the FA instruction set with nearly 2.9 times increase in median task success and dramatically increased average fitness. The fully-associative capability, facilitating specific instruction formats, appears crucial within the highly diverse Logic-77 environment. Indeed, on average 9.2\% of the instructions that may utilize more than one nop-modifier that were present in the dominant genotype at the end of the \instset{FA} experiments with the Logic-77 environment indeed used more than one nop.   The Fibonacci-32 environment also sees a notable 44\% improvement in task success, with a corresponding increase in fitness. Mean usage of multiple nop modifiers was 16.4\% of multi-nop modifiable instructions in the final dominant genotype of the Fibonacci runs. The Sort-10 and Match-12 environments show statistically significant gains for both metrics, but none of these improvements are substantial in nature.  The Navigation environment shows a slight, non-significant decline in fitness ($p < 0.054$, Wilcoxon rank-sum test) and task success ($p < 0.178$, Wilcoxon rank-sum test) when tested with the FA instruction set.

\subsection*{Number of Registers}
The \instset{Register}-series instruction sets generally show little variation in performance (Tables \ref{tbl:is:r:fit} and \ref{tbl:is:r:ts}). In the Logic-77 environment there is a slight positive trend as the number of registers increases, but none are significant after Bonferroni correction, and the magnitudes of the changes are not particularly notable.  The only substantial differences observed among all tested configurations are a drop in task success and a drop in fitness with \instset{R16} in the Logic-9 environment, indicating a potential drag on the system due to the dramatic increase in instruction set size with the addition of 13 more nops, though not as severe as completely non-functional bloat \cite{BrysonOfria2012}.  The Sort-10 environment demonstrates significant loss of performance in all treatments, relative to the \instset{FA} architecture, though none of the variation observed is substantial in nature ($\ll$ 1\% difference in task success). The Navigation environment does show what initially appears to be a substantial uptick in performance under \instset{R16}, but with task success still well below 1\%, it is not enough to allow the populations to complete this task. It does, however, indicate that we may wish to explore higher register counts again in configurations where populations have more success with this task.

\subsection*{Explicit Labels}
The \instset{Label}-series instruction sets show mixed results (Tables \ref{tbl:is:label:fit} and \ref{tbl:is:label:ts}). The Logic-9, Limited-9, Sort-10, and Navigation environments show virtually no substantial differences in task success, regardless of the set used.  The Limited-9, Sort-10, and Navigation environments shows slight positive fitness trends as more labeling options are included in the instruction set. The Logic-77 environment shows significantly detrimental results for both fitness and task success when only the label-based instructions are included. When any form of sequence matching instructions are included in the Logic-77 environment, both metrics return to the reference levels. The Match-12 environment shows no significant difference for either metric among all but one instruction set. \instset{Label-Seq-Both}, the most complete instruction set in this group, shows a notably significant drop of both metrics in the Match environment.  The Fibonacci-32 environment shows positive gains in all \instset{Label}-series instruction sets. The positive gains observed in the Fibonacci-32 environment were both significant and substantial, with 24.6\% and 27.2\% improvement in fitness and task success, respectively, when using the \instset{Label-Seq-Both} instruction set. In the Navigation environment using the \instset{Label-Seq-Direct} instruction set, 8 outlier populations notably demonstrated task success greater than 0.10, with two at 0.141, indicating that substantial progress was made in those particular runs. No previous runs in this environment have exhibited such success in the short time period of 100,000 updates used \cite{GrabowskiBrysonPennock2010,Grabowski:2011ur}. 

\subsection*{Split Input/Output Operations}

The \instset{Split-IO} instruction set shows improvements that are both significant and often substantial in the Logic-9 and Logic-77 environments, the Match-12 environment, and the Fibonacci-32 environment (Tables \ref{tbl:is:splitio:fit} and \ref{tbl:is:splitio:ts}). Indeed the Logic-77 and Fibonacci-32 environments show 21\% and 17\% improvements in median task success, respectively. The Sort-10 environment, on the other hand, completely collapses, showing effectively 0 task success and correspondingly low fitness. The Limited-9 environment shows mixed results, with a small gain in task success but a drop in fitness. The Navigation environment shows marginal drops in both metrics, though neither significant and, similar to previous instruction sets tested, still well below 1\% of the success possible.

\subsection*{Search}

The three \instset{Search}-series instruction sets showed little measurable difference in performance for the Logic-9, Match-12, Fibonacci-32, Sort-10, Limited-9, and Navigation environments (Tables \ref{tbl:is:search:fit} and \ref{tbl:is:search:ts}). The Logic-77 environment showed small, significant drops in fitness for all sets, with a corresponding drop in task success.

In the \instset{Search-Goto} instruction set, we initially tested a variant of the \instr{jmphead} instruction, which changed the default head it operated on to be the \flowhead. A notable and often significant drop in fitness was observed in all seven environments with these two instruction sets, leading to the architectures explored here.

\subsection*{Flow Control}
The \instset{Flow}-series instruction sets tested three groups of flow control instructions separately and in several combinations (Tables \ref{tbl:is:flow:fit} and \ref{tbl:is:flow:ts}). Throughout all instruction sets tested, the Fibonacci-32 environment showed no significant variation from the \instset{Search-Directional} instruction set performance. The Match-12 environment had some significant drops in fitness, but these were not substantial and also not coupled with a drop in task success. The Logic-9 environment showed significant, though again insubstantial, loss of fitness with all \instset{Flow}-series instruction sets. Three instruction sets, \instset{Flow-If0}, \instset{Flow-IfX}, and \instset{Flow-If0-IfX-MovHead}, had corresponding small significant decreases in task success.

Individually, the \instset{If0} instruction group made virtually no difference in performance among any of the seven environments. When tested in combination with the other instruction groups, there is no clear indication of interaction, positive or negative.

The \instset{IfX} instruction group both individually and in combination with other groups shows positive gains in the Navigation environment, both fitness and task success. This outcome is likely due to the nature of the signposts in this environment \cite{GrabowskiBrysonPennock2010}, such that comparing against certain ``magic'' numbers for decision making is likely beneficial. The remaining six environments show no substantial variation attributable to these instructions.

\begin{figure}[!ht]
\begin{center}
\includegraphics{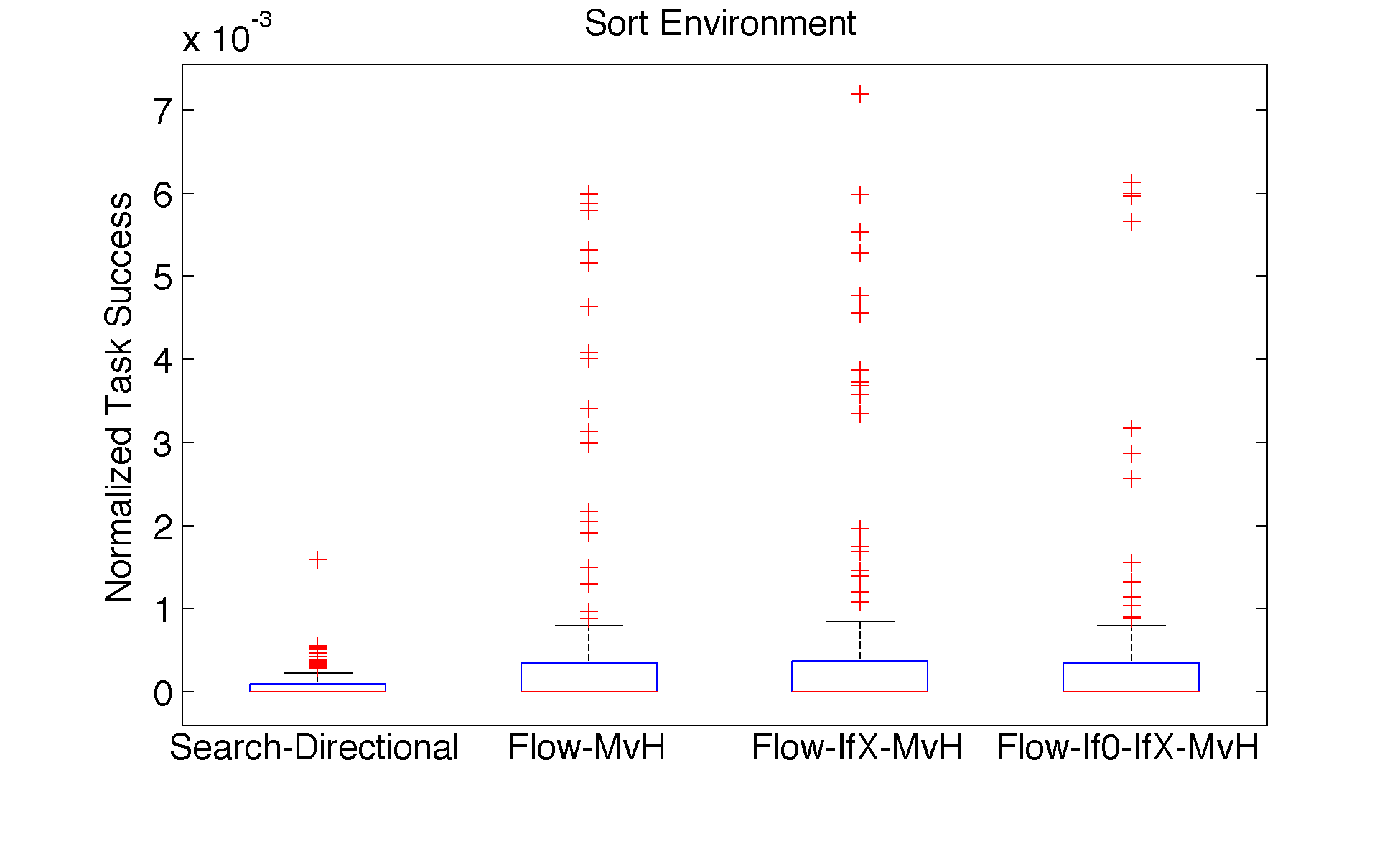}
\caption{\textbf{Normalized task success distributions of selected \instset{Flow}-series instruction sets in the Sort-10 environment.}}
\label{fig:is:flowsort}
\end{center}
\end{figure}

The third instruction group, \instset{MovHead}, shows the greatest variation in performance among those tested.  In the Logic-77 environment, all instruction sets containing the \instset{MovHead} group show substantial decreases in median fitness, 14.3\% on average.  The two combination sets containing \instset{MovHead}, \instset{Flow-IfX-MovHead} and \instset{Flow-If0-IfX-MovHead}, also show corresponding decreases in task success in the Logic-77 environment. The Sort-10, Limited-9, and Navigation environments, on the other hand, show substantial improvements in task success, and often fitness, for all three instruction sets containing the \instset{MovHead} group. The Navigation environment, notably, approaches median task success around 1\% when the \instset{IfX} and \instset{MovHead} instruction groups are combined, indicating the importance of effective flow control for that environment. The Sort-10 environment improvements are difficult to observe from median values.  Indeed the greatest driver of the improvements are infrequent outliers approaching 0.7\% task success, the highest ever observed in the Sort-10 environment (see Figure \ref{fig:is:flowsort}).

\section*{Discussion}

\begin{figure}[!ht]
\begin{center}
\includegraphics[width=6.5in]{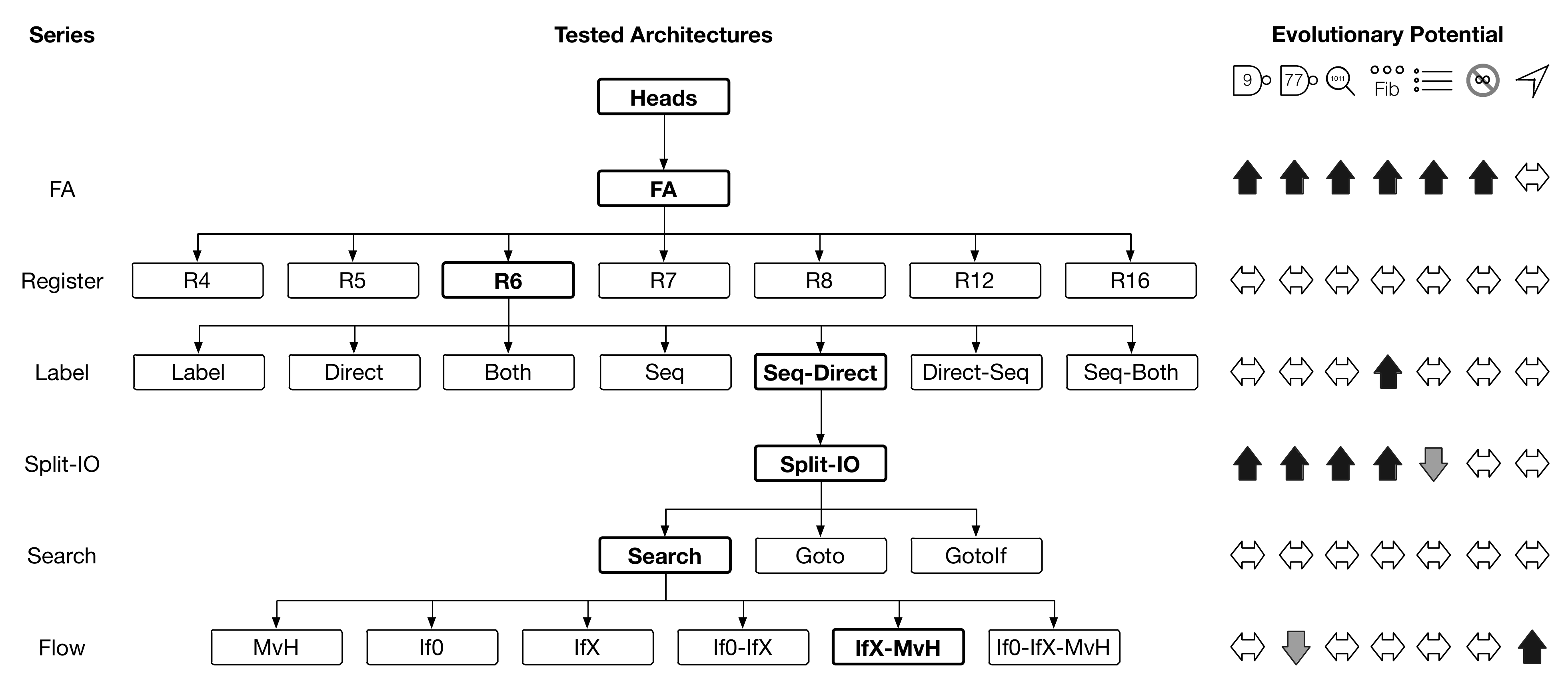}
\caption{\textbf{The order and relationship of all tested architecture modifications (center), organized by instruction set series (left). The evolutionary potential of the architecture selected as the basis for further experiments in each series (shown in bold) is displayed (right) for the Logic-9, Logic-77, Match-12, Fibonacci-32, Sort-10, Limited-9, and Navigation environments, respectively. Up arrows (black) indicate increased potential, down arrows (gray) indicate decreased potential, and double ended arrows (white) denote no significant trend. In general, FA (fully-associative) and Split-IO (separated input and output operations) demonstrated broadly beneficial impacts on evolutionary potential. The remaining tested modifications highlight the robustness of digital evolution, exhibiting no systematic effects on evolutionary potential. }}
\label{fig:summary}
\end{center}
\end{figure}

We have investigated the evolutionary potential of six groups of modified instruction set architectures of a digital evolution system, each within seven different computational environments (see Figure \ref{fig:summary}). Among the groups investigated there were three classes of outcomes, broad multi-environment improvement, mixed results, and no discernible trend. Notably absent from the observed classes were changes that were negative on balance, let alone broadly detrimental; although, this was not entirely unexpected since the particular changes we chose to test were ones that we expected could help. Some instruction set architectures did demonstrate decreased performance in the mixed result grouping, yet only one example demonstrated highly substantial degradation, the \instset{Split-IO} instruction set in the Sort-10 environment. We explore potential explanations for this particular case below. \textbf{In general, evolution has proven to be surprisingly robust to the explored genetic hardware changes, regardless of environment.}

Two groups of instruction-set modifications yielded broadly beneficial changes in both fitness and task success.  The \instset{Fully-Associative} (\instset{FA}) architectures instruction data flow enhancements led to highly significant gains in five of the seven environments.  The remaining two environments, Sort-10 and Navigation, show some slight improvement and no discernible difference, respectively.  The second group that demonstrated broadly positive results was the \instset{Split-IO} instruction set.  The separation of the input and output operation allows finer-grained data flow between the CPU and the environment.  This control afforded by the \instset{Split-IO} architecture was beneficial to the same five environments as the \instset{FA} architecture.  The Navigation environment showed no particular change in fitness performance, and a small, but insubstantial change in task success.  The only major detriment to the splitting of input and output operations was observed in the Sort-10 environment. As a whole, these two groups indicate that it is beneficial to maintain as much flexibility as possible with regard to instruction interactions. This flexibility allows evolution to finely tune interactions, yielding greater evolutionary potential.

The \instset{Register}-series, \instset{Label}-series, and \instset{Search}-series architectures all demonstrated no discernible trend in performance, despite representing 17 of the 25 tested architectures. There were some particular environment/instruction set combinations that had significant variations, yet these were rarely substantial in nature. It is particularly surprising that the \instset{Register}-series instruction sets showed such minimal deviation, given that going from the \instset{FA} architecture to the \instset{R16} architecture represents a greater than five-fold increase in working set and a 50\% increase in instruction set size. Similarly, the \instset{Label-Seq-Both} instruction set represents a 20.6\% increase in instruction set size, with no substantially negative effect. Taken together these groups provide additional evidence that the evolutionary process is rather robust to genetic language dilution \cite{BrysonOfria2012}, maintaining the ability to adapt successfully to the environment despite searching a much larger genotype space.

The \instset{Flow}-series of instruction set architectures represents a third class of outcomes, yielding improved results in a subset of environments and degradation of performance in one environment. The Sort-10, Limited-9, and Navigation environments all show substantial gains in both fitness and task success metrics when using instruction sets containing the \instset{IfX} and \instset{MovHead} instruction groups. The Logic-77 environment, on the other hand, shows a notable drop in performance. It is possible that this environment does not require a great deal of flow control, thus is being negatively affected by the disruptive nature of the additional flow control instructions.  In environments where flow control decisions are critical for success, such as the Sort-10 and Navigation environment, the benefits of more flexible flow control outweigh their disruptive effects.

The Sort-10 environment stands out as the only example where a single, small change -- splitting the input and output instructions -- made a large destructive difference in performance. Median task success collapsed to be statistically indistinguishable from 0, and remained there despite further beneficial instruction set modifications. These results are likely an artifact of the environment itself, rather than a general trend. We set up the Sort-10 environment to control for random inputs and to, on average, provide no benefit unless active sorting was performed by an organism. However, the inputs for sorting are indeed a random sample of 10 integers. It is possible, due to chance, for a partial ordering of numbers to yield a positive metabolic reward even if the sequence of inputs is simply echoed back to the environment. When using instruction sets featuring the paired-input-and-output instruction, simply mutating this instruction into the section of the genome responsible for replication may be enough to confer the echo capability, presenting an opportunity for lucky organisms to occasionally reap rewards. When the operations are split into two separate instructions, it then requires two coordinated mutations to confer the echo capability and doubles the execution cost for performing the task. The combination of these factors most likely contributes to the observed drop in median performance.

Instruction data flow, working set size, and flow control are the three main features addressed by the six groups of instruction set modifications presented here. All of these features play an important role in implementing a successful sorting algorithm. Despite the modifications in the instruction set architectures we tested, no significantly beneficial change was observed in either fitness or task success within the Sort-10 environment. Most likely, the highly constrained memory size of these architectures limits the potential within this environment. In fact, a hand-written organism that performs the task successfully with the \instset{Heads} architecture requires nearly every single stack location in both available stacks. Another factor limiting potential may simply be the time allotted for evolution, which was held constant in our current study. The additional flow control instructions tested in the \instset{Flow}-series architectures show some signs of improved success in this environment, with numerous outlier populations.  Given additional time to evolve, these and other populations would likely be able to refine the emerging solutions.

When features from all six instruction set groups are combined to form the \instset{Heads-EX} architecture, significant and substantial improvements relative to the base \instset{Heads} architecture are observed in six of the seven environments (Tables \ref{tbl:is:final:fit} and \ref{tbl:is:final:ts}). Despite these improvements, there still remains a great deal of unexploited opportunity in five of the environments. Specific architectural changes to address these environments may yield greater results, such as the addition of an instruction capable of building arbitrary numbers for the Match-12 environment. However, such focused modifications could mask the need for more sweeping changes. Even with significant gains under two instruction sets, the Logic-77 environment still shows room for substantial improvement, as median task success shows populations utilizing less than 55\% of the opportunities present. Even more so, the Sort-10 and Navigation environments exploit less than 1\% of the available potential.

It is clear from this present study that we have just started to identify the most effective genetic hardware for adaptive evolution in digital organisms and there remains room for significant future improvement. Indeed, our current study has focused on modifications within the framework of von Neumann machine code formalisms. We expect that further studies of instruction set architecture enhancements for evolvable systems, both within the limits of von Neumann architectures and the broader range of programming formalisms, will unlock this potential, facilitating advancements in the application of digital evolution and artificial life.

\section*{Acknowledgments}

The authors would like to thank Jeff Barrick, Matt Rupp, Chris \mbox{Strelioff}, Aaron P. Wagner, Bess Walker and the members of the MSU Digital Evolution Laboratory for comments on the manuscript.

\bibliography{references}

\section*{Figure Legends}

\clearpage

\section*{Tables}

\begin{longtable}{@{\extracolsep{\fill}} l l}
\caption{\label{tbl:is:insts}\bf Instruction Glossary}
\\ [0ex]
\hline\hline \\ [-2.3ex]
Instruction & Description \\ [0.1ex]
\hline \\ [-2.3ex]
\endfirsthead
\hline\hline \\ [-2.3ex]
Instruction & Description \\ [0.1ex]
\hline \\ [-2.3ex]
\endhead
\hline \\
\endfoot
\hline \\
\multicolumn{2}{l}{\begin{minipage}[t]{\columnwidth}Description of the instructions used across all tested instruction set architectures. A register name (AX, BX, CX, etc.) or head (IP, FLOW, etc.) surrounded by question marks refers to the default argument used when executed, subject to nop modification. Instructions depicted in \textbf{bold} are in the default \instset{Heads} instruction set.\end{minipage}}
\endlastfoot
\instrb{add} & Add ?BX? to ?CX? and place the result in ?BX? \\ [.5ex]
\instrb{dec} & Decrement ?BX? by one \\ [.5ex]
\instrb{get-head} & Copy the position of the ?IP? head into ?CX? \\ [.5ex]
\instr{goto} & Move IP to direct match label \\ [.5ex]
\instr{goto-if-n-equ} & Move IP to direct match label if BX != CX \\ [.5ex]
\instr{goto-if-less} & Move IP to direct match label if BX $<$ CX \\ [.5ex]
\instrb{h-alloc} & Allocate maximum allowed space for offspring \\ [.5ex]
\instrb{h-copy} & Copy from read-head to write-head; advance both \\ [.5ex]
\instrb{h-divide} & Divide code between read and write heads as offspring \\ [.5ex]
\instrb{if-copied-seq-comp} & Execute next instruction if just copied complement sequence \\ [.5ex]
\instr{if-copied-seq-direct} & Execute next instruction if just copied direct-match sequence \\ [.5ex]
\instr{if-copied-lbl-comp} & Execute next instruction if just copied complement label \\ [.5ex]
\instr{if-copied-lbl-direct} & Execute next instruction if just copied direct-match label \\ [.5ex]
\instr{if-equ-0} & Execute next instruction if ?BX? = 0, else skip it \\ [.5ex]
\instr{if-equ-x} & Execute next instruction if BX = ?nop-defined constant? , else skip it \\ [.5ex]
\instr{if-gtr-0} & Execute next instruction if ?BX? $>$ 0, else skip it \\ [.5ex]
\instr{if-gtr-x} & Execute next instruction if BX $>$ ?nop-defined constant? , else skip it  \\ [.5ex]
\instrb{if-less} & Execute next instruction if ?BX? $<$ ?CX?, else skip it \\ [.5ex]
\instr{if-less-0} & Execute next instruction if ?BX? $<$ 0, else skip it \\ [.5ex]
\instrb{if-n-equ} & Execute next instruction if ?BX? != ?CX?, else skip it \\ [.5ex]
\instr{if-not-0} & Execute next instruction if ?BX? != 0, else skip it \\ [.5ex]
\instrb{inc} & Increment ?BX? by one \\ [.5ex]
\instr{input} & Input new number into ?BX? \\ [.5ex]
\instrb{IO} & Output ?BX?, and input new number back into ?BX? \\ [.5ex]
\instrb{jmp-head} & Move head ?Flow? by amount in ?CX? register \\ [.5ex]
\instr{label} & No-operation; marks the beginning of a genome position label \\ [.5ex]
\instrb{mov-head} & Move head ?IP? to the flow head \\ [.5ex]
\instr{mov-head-if-less} & Move head ?IP? to the flow head if ?BX? $<$ ?CX? \\ [.5ex]
\instr{mov-head-if-n-equ} & Move head ?IP? to the flow head if ?BX? != ?CX? \\ [.5ex]
\instrb{nand} & Nand ?BX? by ?CX? and place the result in ?BX? \\ [.5ex]
\instrb{nop-A} & No-operation; modifies other instructions \\ [.5ex]
\instrb{nop-B} & No-operation; modifies other instructions \\ [.5ex]
\instrb{nop-C} & No-operation; modifies other instructions \\ [.5ex]
\instr{nop-D} & No-operation; modifies other instructions \\ [.5ex]
\instr{nop-E} & No-operation; modifies other instructions \\ [.5ex]
\instr{nop-F} & No-operation; modifies other instructions \\ [.5ex]
\instr{nop-G} & No-operation; modifies other instructions \\ [.5ex]
\instr{nop-H} & No-operation; modifies other instructions \\ [.5ex]
\instr{nop-I} & No-operation; modifies other instructions \\ [.5ex]
\instr{nop-J} & No-operation; modifies other instructions \\ [.5ex]
\instr{nop-K} & No-operation; modifies other instructions \\ [.5ex]
\instr{nop-L} & No-operation; modifies other instructions \\ [.5ex]
\instr{nop-M} & No-operation; modifies other instructions \\ [.5ex]
\instr{nop-N} & No-operation; modifies other instructions \\ [.5ex]
\instr{nop-O} & No-operation; modifies other instructions \\ [.5ex]
\instr{nop-P} & No-operation; modifies other instructions \\ [.5ex]
\instr{output} & Output ?BX? \\ [.5ex]
\instrb{pop} & Remove top number from stack and place into ?BX? \\ [.5ex]
\instrb{push} & Copy number from ?BX? and place it into the stack \\ [.5ex]
\instr{search-lbl-comp-s} & Find complement label from genome start and move the flow head \\ [.5ex]
\instr{search-lbl-direct-b} & Find direct label backward and move the flow head \\ [.5ex]
\instr{search-lbl-direct-f} & Find direct label forward and move the flow head \\ [.5ex]
\instr{search-lbl-direct-s} & Find direct label from genome start and move the flow head \\ [.5ex]
\instrb{search-seq-comp-s} & Find complement sequence from genome start and move the flow head \\ [.5ex]
\instr{search-seq-direct-b} & Find direct sequence backward and move the flow head \\ [.5ex]
\instr{search-seq-direct-f} & Find direct sequence forward and move the flow head \\ [.5ex]
\instr{search-seq-direct-s} & Find direct sequence from genome start and move the flow head \\ [.5ex]
\instrb{set-flow} & Set flow-head to position in ?CX? \\ [.5ex]
\instr{sg-move} & Move one location forward in the Navigation environment \\ [.5ex]
\instr{sg-rotate-l} & Rotate heading 45\% left in the Navigation environment \\ [.5ex]
\instr{sg-rotate-r} & Rotate heading 45\% right in the Navigation environment \\ [.5ex]
\instr{sg-sense} & Read the value of the current location in the Navigation environment \\ [.5ex]
\instrb{shift-r} & Shift bits in ?BX? right by one (divide by two) \\ [.5ex]
\instrb{shift-l} & Shift bits in ?BX? left by one (multiply by two) \\ [.5ex]
\instrb{sub} & Subtract ?CX? from ?BX? and place the result in ?BX? \\ [.5ex]
\instrb{swap} & Swap the contents of ?BX? with ?CX? \\ [.5ex]
\instrb{swap-stk} & Toggle which stack is currently being used \\ [.5ex]
\hline\hline
\end{longtable}

\begin{table}[!ht]
\begin{flushleft}
\caption{\label{tbl:is:lblsets}\bf \instset{Label} Instruction Sets Tested}
\end{flushleft}
\begin{tabular*}{\hsize}{@{\extracolsep{\fill}} l c c c c c c c c c c c c c }
\\ [-2ex]
\textbf{Instruction Set} & \rotatebox{90}{\instr{label}} & \rotatebox{90}{\instr{if-copied-lbl-comp}} & \rotatebox{90}{\instr{if-copied-lbl-direct}} & \rotatebox{90}{\instr{if-copied-seq-comp}} & \rotatebox{90}{\instr{if-copied-seq-direct}} & \rotatebox{90}{\instr{search-lbl-comp-s}} & \rotatebox{90}{\instr{search-lbl-direct-s}} & \rotatebox{90}{\instr{search-seq-comp-s}} & \rotatebox{90}{\instr{search-seq-direct-s}} \\ [0.1ex]
\hline \\ [-2.3ex]
\instset{R6}               &             &             &             & \textbullet &             &             &             & \textbullet &             \\ [.5ex]
\instset{Label}            & \textbullet & \textbullet &             &             &             & \textbullet &             &             &             \\ [.5ex]
\instset{Label-Direct}     & \textbullet &             & \textbullet &             &             &             & \textbullet &             &             \\ [.5ex]
\instset{Label-Both}       & \textbullet & \textbullet & \textbullet &             &             & \textbullet & \textbullet &             &             \\ [.5ex]
\instset{Label-Seq}        & \textbullet & \textbullet &             & \textbullet &             & \textbullet &             & \textbullet &             \\ [.5ex]
\instset{Label-Seq-Direct} & \textbullet &             & \textbullet &             & \textbullet &             & \textbullet &             & \textbullet \\ [.5ex]
\instset{Label-Direct-Seq} & \textbullet &             & \textbullet & \textbullet &             &             & \textbullet & \textbullet &             \\ [.5ex]
\instset{Label-Seq-Both}   & \textbullet & \textbullet & \textbullet & \textbullet & \textbullet & \textbullet & \textbullet & \textbullet & \textbullet \\ [.5ex]
\hline\hline
\end{tabular*}
\begin{flushleft}
Marks in each column indicating that the set contains the relevant instruction.
\end{flushleft}
\end{table}

\begin{table}[!ht]
\begin{flushleft}
\caption{\label{tbl:is:flowsets}\bf \instset{Flow} Instruction Sets Tested}
\end{flushleft}
\begin{tabular*}{\hsize}{@{\extracolsep{\fill}} l c c c}
\\ [-2ex]
\hline\hline \\ [-2.3ex]
\textbf{Instruction Set} & \instset{If0} Instructions & \instset{IfX} Instructions & \instset{MovHead} Instructions \\ [0.1ex]
\hline \\ [-2.3ex]
\instset{Flow-If0} & \textbullet & & \\ [.5ex]
\instset{Flow-IfX} & & \textbullet & \\ [.5ex]
\instset{Flow-MvH} & & & \textbullet \\ [.5ex]
\instset{Flow-If0-MvH} & \textbullet & \textbullet & \\ [.5ex]
\instset{Flow-IfX-MvH} & & \textbullet & \textbullet \\ [.5ex]
\instset{Flow-If0-IfX-MvH} & \textbullet & \textbullet & \textbullet \\ [.5ex]
\hline\hline
\end{tabular*}
\begin{flushleft}
Instruction set by row, with marks in each column indicating that the set contains the relevant instruction group.
\end{flushleft}
\end{table}

\begin{table}[!ht]
\begin{flushleft}
\caption{\label{tbl:is:fa:fit}\bf \instset{Heads} and \instset{Fully-Associative} Architectures Fitness}
\end{flushleft}
\begin{tabular*}{\hsize}{@{\extracolsep{\fill}} l l l l l l l l}
\\ [-2ex]
\hline\hline \\ [-2.3ex]
& Logic-9 & Logic-77 & Match-12 & Fibonacci-32 & Sort-10 & Limited-9 & Navigation \\ [0.1ex]
\hline \\ [-2.3ex]
\multirow{2}{*}{\instset{Heads}} & 19.07 & 12.43 & 0.173 & 3.730 & -0.54 & 4.430 & 1.071 \\ 
 & \scriptsize{(17.71, 19.76)} & \scriptsize{(11.51, 14.22)} &  \scriptsize{(0.146, 0.224)} & \scriptsize{(3.300, 4.050)} & \scriptsize{(-0.63, -0.45)} & \scriptsize{(4.283, 4.595)} & \scriptsize{(1.035, 1.383)} \\ [.5ex]
\hline \\ [-2.3ex]
\multirow{2}{*}{\instset{FA}} & \textbf{22.99} & \textbf{39.35} & \textbf{0.215} & \textbf{4.806} & \textbf{-0.38} & \textbf{4.840} & 1.038 \\ 
 & \scriptsize{(22.70, 23.08)} & \scriptsize{(35.05, 41.83)} &  \scriptsize{(0.191, 0.251)} & \scriptsize{(4.474, 5.212)} & \scriptsize{(-0.45, -0.33)} & \scriptsize{(4.671, 5.082)} & \scriptsize{(1.022, 1.069)} \\ [.5ex]
\hline\hline
\end{tabular*} \\ [0.5ex]
\begin{flushleft}
Fitness results of the \instset{Heads} and \instset{Fully-Associative} (\instset{FA}) instruction set architectures, where multiple nop arguments can modify the behavior of an instruction. Each entry shows the median log$_2$ population mean fitness in the respective environment, with $95\%$ confidence intervals in parentheses. Bold entries indicate significant ($p < 0.05$, Wilcoxon rank-sum test) deviations after sequential Bonferroni correction.
\end{flushleft}
\end{table}

\begin{table}[!ht]
\begin{flushleft}
\caption{\label{tbl:is:fa:ts}\bf \instset{Heads} and \instset{Fully-Associative} Architectures Task Success}
\end{flushleft}
\begin{tabular*}{\hsize}{@{\extracolsep{\fill}} l l l l l l l l}
\\ [-2ex]
\hline\hline \\ [-2.3ex]
& Logic-9 & Logic-77 & Match-12 & Fib.-32 & Sort-10 & Limited-9 & Navigation \\ [0.1ex]
\hline \\ [-2.3ex]
\multirow{2}{*}{\instset{Heads}} & 0.829 & 0.176 & 0.145 & 0.206 & 1.31$\times10^{-4}$ & 0.909 & 3.97$\times10^{-3}$ \\ 
 & \scriptsize{(0.752, 0.839)} & \scriptsize{(0.161, 0.198)} & \scriptsize{(0.145, 0.146)} & \scriptsize{(0.178, 0.238)} & \scriptsize{(1.08, 1.47)} & \scriptsize{(0.894, 0.913)} & \scriptsize{(3.96, 4.35)} \\ [.5ex]
\hline \\ [-2.3ex]
\multirow{2}{*}{\instset{FA}} & \textbf{0.936} & \textbf{0.505} & \textbf{0.148} & \textbf{0.297} & 1.55$\times10^{-4}$ & \textbf{0.927} & 3.96$\times10^{-3}$ \\ 
 & \scriptsize{(0.930, 0.943)} & \scriptsize{(0.453, 0.546)} & \scriptsize{(0.147, 0.149)} & \scriptsize{(0.278, 0.332)} & \scriptsize{(1.44, 1.67)} & \scriptsize{(0.924, 0.929)} & \scriptsize{(3.95, 3.97)} \\ [.5ex]
\hline\hline
\end{tabular*} \\ [0.5ex]
\begin{flushleft}
Task success results of the \instset{Heads} and \instset{Fully-Associative} (\instset{FA}) instruction set architectures.  Each entry shows the median normalized task success in the respective environment, with $95\%$ confidence intervals in parentheses. Bold entries denote significant ($p < 0.05$, Wilcoxon rank-sum test) deviations.
\end{flushleft}
\end{table}

\begin{table}[!ht]
\begin{flushleft}
\caption{\label{tbl:is:r:fit}\bf \instset{Register} Series Architectures Fitness}
\end{flushleft}
\begin{tabular*}{\hsize}{@{\extracolsep{\fill}} l l l l l l l l}
\\ [-2ex]
\hline\hline \\ [-2.3ex]
& Logic-9 & Logic-77 & Match-12 & Fib.-32 & Sort-10 & Limited-9 & Navigation \\ [0.1ex]
\hline \\ [-2.3ex]
\multirow{2}{*}{\instset{FA}} & 22.63 & 38.85 & 0.223 & 4.828 & 0.09 & 4.934 & 1.027 \\ 
 & \scriptsize{(22.39, 23.01)} & \scriptsize{(34.76, 43.67)} &  \scriptsize{(0.191, 0.273)} & \scriptsize{(4.255, 5.332)} & \scriptsize{(-0.07, 0.21)} & \scriptsize{(4.661, 5.282)} & \scriptsize{(1.016, 1.042)} \\ [.5ex]
\hline \\ [-2.3ex]
\multirow{2}{*}{\instset{R4}} & 22.85 & 38.70 & 0.243 & 4.666 & \textbf{-0.39} & 5.253 & \textbf{1.080} \\ 
 & \scriptsize{(22.67, 23.01)} & \scriptsize{(33.90, 43.02)} &  \scriptsize{(0.204, 0.290)} & \scriptsize{(4.233, 5.142)} & \scriptsize{(-0.48, -0.32)} & \scriptsize{(4.925, 5.514)} & \scriptsize{(1.054, 1.792)} \\ [.5ex]
\hline \\ [-2.3ex]
\multirow{2}{*}{\instset{R5}} & 22.73 & 38.42 & 0.231 & 5.067 & \textbf{-0.45} & 5.158 & \textbf{1.083} \\ 
 & \scriptsize{(22.50, 22.86)} & \scriptsize{(34.43, 42.54)} &  \scriptsize{(0.206, 0.281)} & \scriptsize{(4.540, 5.623)} & \scriptsize{(-0.57, -0.39)} & \scriptsize{(4.300, 5.400)} & \scriptsize{(1.056, 1.340)} \\ [.5ex]
\hline \\ [-2.3ex]
\multirow{2}{*}{\instset{R6}} & 22.78 & 43.01 & 0.229 & 4.908 & \textbf{-0.43} & 5.117 & \textbf{1.111} \\ 
 & \scriptsize{(22.29, 22.97)} & \scriptsize{(40.01, 45.90)} &  \scriptsize{(0.206, 0.274)} & \scriptsize{(4.293, 5.719)} & \scriptsize{(-0.54, -0.34)} & \scriptsize{(4.925, 5.374)} & \scriptsize{(1.080, 2.730)} \\ [.5ex]
\hline \\ [-2.3ex]
\multirow{2}{*}{\instset{R7}} & 22.75 & 43.41 & 0.204 & 4.598 & \textbf{-0.40} & 5.135 & \textbf{1.562} \\ 
 & \scriptsize{(22.58, 22.97)} & \scriptsize{(38.97, 45.67)} &  \scriptsize{(0.177, 0.225)} & \scriptsize{(4.174, 5.078)} & \scriptsize{(-0.49, -0.29)} & \scriptsize{(4.978, 5.407)} & \scriptsize{(1.096, 3.234)} \\ [.5ex]
\hline \\ [-2.3ex]
\multirow{2}{*}{\instset{R8}} & 22.75 & 43.04 & \textbf{-0.027} & 4.831 & \textbf{-0.47} & 5.292 & \textbf{1.156} \\ 
 & \scriptsize{(22.55, 22.95)} & \scriptsize{(39.25, 47.78)} &  \scriptsize{(-0.07, 0.19)}  & \scriptsize{(4.392, 5.308)} & \scriptsize{(-0.57, -0.33)} & \scriptsize{(5.058, 5.736)} & \scriptsize{(1.099, 2.815)} \\ [.5ex]
\hline \\ [-2.3ex]
\multirow{2}{*}{\instset{R12}} & 22.62 & 44.26 & \textbf{-0.11} & 4.678 & \textbf{-0.54} & 5.180 & \textbf{1.377} \\ 
 & \scriptsize{(22.45, 22.76)} & \scriptsize{(40.82, 48.18)} &  \scriptsize{(-0.12, -0.08)} & \scriptsize{(4.082, 5.244)} & \scriptsize{(-0.56, -0.49)} & \scriptsize{(4.901, 5.621)} & \scriptsize{(1.114, 3.012)} \\ [.5ex]
\hline \\ [-2.3ex]
\multirow{2}{*}{\instset{R16}} & \textbf{21.68} & 42.26 & \textbf{-0.11} & 4.028 & \textbf{-0.55} & \textbf{5.734} & \textbf{3.78} \\ 
 & \scriptsize{(19.76, 22.22)} & \scriptsize{(40.02, 46.26)} &  \scriptsize{(-0.13, -0.10)} & \scriptsize{(3.620, 4.474)} & \scriptsize{(-0.59, -0.50)} & \scriptsize{(5.390, 6.466)} & \scriptsize{(1.157, 3.326)} \\ [.5ex]
\hline\hline
\end{tabular*} \\ [0.5ex]
\begin{flushleft}
Fitness results of the \instset{Register}-series instruction set architectures, which vary the number of registers available in the virtual CPUs. Each entry shows the median log$_2$ population mean fitness in the respective environment, with $95\%$ confidence intervals in parentheses. Bold entries indicate significant ($p < 0.05$, Wilcoxon rank-sum test) deviations after sequential Bonferroni correction.
\end{flushleft}
\end{table}

\begin{table}[!ht]
\begin{flushleft}
\caption{\label{tbl:is:r:ts}\bf \instset{Register} Series Architectures Task Success}
\end{flushleft}
\begin{tabular*}{\hsize}{@{\extracolsep{\fill}} l l l l l l l l}
\\ [-2ex]
\hline\hline \\ [-2.3ex]
& Logic-9 & Logic-77 & Match-12 & Fib.-32 & Sort-10 & Limited-9 & Navigation \\ [0.1ex]
\hline \\ [-2.3ex]
\multirow{2}{*}{\instset{FA}} & 0.932 & 0.495 & 0.147 & 0.288 & 2.53$\times10^{-4}$ & 0.926 & 3.96$\times10^{-3}$ \\ 
 & \scriptsize{(0.921, 0.938)} & \scriptsize{(0.452, 0.565)} & \scriptsize{(0.146, 0.148)} & \scriptsize{(0.263, 0.307)} & \scriptsize{(2.21, 2.74)} & \scriptsize{(0.923, 0.929)} & \scriptsize{(3.95, 3.96)} \\ [.5ex]
\hline \\ [-2.3ex]
\multirow{2}{*}{\instset{R4}} & 0.937 & 0.506 & 0.146 & 0.276 & \textbf{1.52$\mathbf{\times10^{-4}}$} & 0.923 & \textbf{3.97$\mathbf{\times10^{-3}}$} \\ 
 & \scriptsize{(0.929, 0.941)} & \scriptsize{(0.441, 0.554)} & \scriptsize{(0.145, 0.148)} & \scriptsize{(0.256, 0.289)} & \scriptsize{(1.42, 1.67)} & \scriptsize{(0.920, 0.929)} & \scriptsize{(3.96, 5.05)} \\ [.5ex]
\hline \\ [-2.3ex]
\multirow{2}{*}{\instset{R5}} & 0.936 & 0.493 & \textbf{0.145} & 0.300 & \textbf{1.38$\mathbf{\times10^{-4}}$} & 0.927 & \textbf{3.97$\mathbf{\times10^{-3}}$} \\ 
 & \scriptsize{(0.929, 0.940)} & \scriptsize{(0.450, 0.544)} & \scriptsize{(0.144, 0.147)} & \scriptsize{(0.284, 0.327)} & \scriptsize{(1.09, 1.62)} & \scriptsize{(0.923, 0.929)} & \scriptsize{(3.96, 4.20)} \\ [.5ex]
\hline \\ [-2.3ex]
\multirow{2}{*}{\instset{R6}} & 0.932 & 0.563 & \textbf{0.145} & 0.294 & \textbf{1.49$\mathbf{\times10^{-4}}$} & 0.930 & \textbf{3.98$\mathbf{\times10^{-3}}$} \\ 
 & \scriptsize{(0.927, 0.940)} & \scriptsize{(0.521, 0.592)} & \scriptsize{(0.144, 0.147)} & \scriptsize{(0.268, 0.326)} & \scriptsize{(1.14, 1.160)} & \scriptsize{(0.926, 0.932)} & \scriptsize{(3.97, 6.68)} \\ [.5ex]
\hline \\ [-2.3ex]
\multirow{2}{*}{\instset{R7}} & 0.940 & 0.554 & \textbf{0.144} & 0.281 & \textbf{1.51$\mathbf{\times10^{-4}}$} & 0.928 & \textbf{4.47$\mathbf{\times10^{-3}}$} \\ 
 & \scriptsize{(0.930, 0.943)} & \scriptsize{(0.502, 0.592)} & \scriptsize{(0.142, 0.146)} & \scriptsize{(0.247, 0.305)} & \scriptsize{(1.26, 1.63)} & \scriptsize{(0.923, 0.932)} & \scriptsize{(3.98, 7.65)} \\ [.5ex]
\hline \\ [-2.3ex]
\multirow{2}{*}{\instset{R8}} & 0.938 & 0.555 & \textbf{0.079} & 0.299 & \textbf{1.33$\mathbf{\times10^{-4}}$} & 0.927 & \textbf{3.99$\mathbf{\times10^{-3}}$} \\ 
 & \scriptsize{(0.931, 0.942)} & \scriptsize{(0.504, 0.613)} & \scriptsize{(0.078, 0.143)} & \scriptsize{(0.275, 0.323)} & \scriptsize{(1.06, 1.57)} & \scriptsize{(0.924, 0.929)} & \scriptsize{(3.97, 6.19)} \\ [.5ex]
\hline \\ [-2.3ex]
\multirow{2}{*}{\instset{R12}} & 0.939 & 0.575 & \textbf{0.078} & 0.298 & \textbf{1.06$\mathbf{\times10^{-4}}$} & 0.930 & \textbf{4.31$\mathbf{\times10^{-3}}$} \\ 
 & \scriptsize{(0.933, 0.943)} & \scriptsize{(0.525, 0.613)} & \scriptsize{(0.077, 0.078)} & \scriptsize{(0.268, 0.318)} & \scriptsize{(1.03, 1.11)} & \scriptsize{(0.928, 0.933)} & \scriptsize{(3.98, 7.33)} \\ [.5ex]
\hline \\ [-2.3ex]
\multirow{2}{*}{\instset{R16}} & 0.910 & 0.550 & \textbf{0.077} & 0.269 & \textbf{1.05$\mathbf{\times10^{-4}}$} & 0.928 & \textbf{7.31$\mathbf{\times10^{-3}}$} \\ 
 & \scriptsize{(0.854, 0.931)} & \scriptsize{(0.524, 0.589)} & \scriptsize{(0.077, 0.078)} & \scriptsize{(0.237, 0.302)} & \scriptsize{(1.01, 1.08)} & \scriptsize{(0.925, 0.932)} & \scriptsize{(3.99, 7.81)} \\ [.5ex]
\hline\hline
\end{tabular*} \\ [0.5ex]
\begin{flushleft}
Task success results of the \instset{Register}-series instruction set architectures.  Each entry shows the median normalized task success in the respective environment, with $95\%$ confidence intervals in parentheses. Bold entries denote significant ($p < 0.05$, Wilcoxon rank-sum test) deviations.
\end{flushleft}
\end{table}

\begin{table}[!ht]
\begin{flushleft}
\caption{\label{tbl:is:label:fit}\bf \instset{Label} Series Architectures Fitness}
\end{flushleft}
\begin{tabular*}{\hsize}{@{\extracolsep{\fill}} l l l l l l l l}
\\ [-2ex]
\hline\hline \\ [-2.3ex]
& Logic-9 & Logic-77 & Match-12 & Fib.-32 & Sort-10 & Limited-9 & Navigation \\ [0.1ex]
\hline \\ [-2.3ex]
\multirow{2}{*}{\instset{R6}} & 22.59 & 38.42 & 0.216 & 4.572 & -0.22 & 5.205 & 1.537 \\ 
 & \scriptsize{(22.10, 22.85)} & \scriptsize{(34.96, 44.12)} &  \scriptsize{(0.201, 0.257)} & \scriptsize{(3.904, 5.134)} & \scriptsize{(-0.32, -0.09)} & \scriptsize{(5.016, 5.518)} & \scriptsize{(1.097, 3.261)} \\ [.5ex]
\hline \\ [-2.3ex]
\multirow{2}{*}{\instset{Label}} & 22.67 & \textbf{28.70} & 0.248 & \textbf{6.213} & -0.42 & 5.430 & 1.926 \\ 
 & \scriptsize{(22.42, 22.94)} & \scriptsize{(25.66, 33.37)} &  \scriptsize{(0.218, 0.316)} & \scriptsize{(5.669, 6.648)} & \scriptsize{(-0.52, -0.33)} & \scriptsize{(5.195, 5.630)} & \scriptsize{(1.108, 3.313)} \\ [.5ex]
\hline \\ [-2.3ex]
\multirow{2}{*}{\instset{Direct}} & 22.50 & \textbf{27.20} & 0.215 & \textbf{5.435} & \textbf{-0.46} & 5.784 & 1.087 \\ 
 & \scriptsize{(21.84, 22.74)} & \scriptsize{(24.38, 31.32)} &  \scriptsize{(0.189, 0.252)} & \scriptsize{(4.545, 6.174)} & \scriptsize{(-0.54, -0.40)} & \scriptsize{(5.429, 6.325)} & \scriptsize{(1.064, 2.742)} \\ [.5ex]
\hline \\ [-2.3ex]
\multirow{2}{*}{\instset{Both}} & 22.32 & \textbf{32.26} & 0.203 & \textbf{6.403} & \textbf{-0.47} & 5.553 & 3.084 \\ 
 & \scriptsize{(19.68, 22.61)} & \scriptsize{(28.96, 38.07)} &  \scriptsize{(0.163, 0.230)} & \scriptsize{(5.715, 6.606)} & \scriptsize{(-0.56, -0.38)} & \scriptsize{(5.098, 5.885)} & \scriptsize{(1.148, 3.396)} \\ [.5ex]
\hline \\ [-2.3ex]
\multirow{2}{*}{\instset{Seq}} & 22.43 & 40.26 & 0.207 & \textbf{6.357} & -0.29 & 5.438 & 2.203 \\ 
 & \scriptsize{(21.94, 22.66)} & \scriptsize{(35.12, 44.36)} &  \scriptsize{(0.134, 0.257)} & \scriptsize{(5.797, 6.733)} & \scriptsize{(-0.36, -0.15)} & \scriptsize{(5.079, 5.755)} & \scriptsize{(1.089, 3.161)} \\ [.5ex]
\hline \\ [-2.3ex]
\instset{Seq} & 22.46 & 44.13 & 0.198 & \textbf{6.175} & -0.33 & 5.651 & 2.335 \\ 
\instset{Direct} & \scriptsize{(22.23, 22.66)} & \scriptsize{(39.73, 48.52)} &  \scriptsize{(0.126, 0.217)} & \scriptsize{(5.212, 6.531)} & \scriptsize{(-0.44, -0.21)} & \scriptsize{(5.441, 5.967)} & \scriptsize{(1.077, 3.335)} \\ [.5ex]
\hline \\ [-2.3ex]
\instset{Direct} & 22.53 & 41.74 & 0.210 & \textbf{5.933} & -0.40 & 5.528 & 2.319 \\ 
\instset{Seq} & \scriptsize{(22.25, 22.69)} & \scriptsize{(38.58, 44.36)} &  \scriptsize{(0.183, 0.300)} & \scriptsize{(5.135, 6.495)} & \scriptsize{(-0.47, -0.23)} & \scriptsize{(5.528, 5.968)} & \scriptsize{(1.093, 3.235)} \\ [.5ex]
\hline \\ [-2.3ex]
\instset{Seq} & 22.44 & 39.56 & \textbf{-0.094} & \textbf{6.116} & -0.33 & \textbf{5.990} & 3.173 \\ 
\instset{Both} & \scriptsize{(21.75, 22.68)} & \scriptsize{(36.64, 42.72)} &  \scriptsize{(-1.42, 0.088)} & \scriptsize{(5.336, 6.430)} & \scriptsize{(-0.45, -0.17)} & \scriptsize{(5.621, 6.374)} & \scriptsize{(2.955, 3.295)} \\ [.5ex]
\hline\hline
\end{tabular*} \\ [0.5ex]
\begin{flushleft}
Fitness results of the \instset{Label}-series instruction set architectures. Each entry shows the median log$_2$ population mean fitness in the respective environment, with $95\%$ confidence intervals in parentheses. Bold entries indicate significant ($p < 0.05$, Wilcoxon rank-sum test) deviations after sequential Bonferroni correction.
\end{flushleft}
\end{table}

\begin{table}[!ht]
\begin{flushleft}
\caption{\label{tbl:is:label:ts}\bf \instset{Label} Series Architectures Task Success}
\end{flushleft}
\begin{tabular*}{\hsize}{@{\extracolsep{\fill}} l l l l l l l l}
\\ [-2ex]
\hline\hline \\ [-2.3ex]
& Logic-9 & Logic-77 & Match-12 & Fib.-32 & Sort-10 & Limited-9 & Navigation \\ [0.1ex]
\hline \\ [-2.3ex]
\multirow{2}{*}{\instset{R6}} & 0.926 & 0.505 & 0.144 & 0.278 & 1.86$\times10^{-4}$ & 0.930 & 4.33$\times10^{-3}$ \\ 
 & \scriptsize{(0.908, 0.937)} & \scriptsize{(0.461, 0.574)} & \scriptsize{(0.142, 0.145)} & \scriptsize{(0.241, 0.303)} & \scriptsize{(1.62, 2.10)} & \scriptsize{(0.926, 0.932)} & \scriptsize{(3.97, 7.76)} \\ [.5ex]
\hline \\ [-2.3ex]
\multirow{2}{*}{\instset{Label}} & 0.941 & \textbf{0.389} & 0.146 & \textbf{0.380} & \textbf{1.45$\mathbf{\times10^{-4}}$} & 0.932 & 4.58$\times10^{-3}$ \\ 
 & \scriptsize{(0.934, 0.945)} & \scriptsize{(0.352, 0.450)} & \scriptsize{(0.144, 0.147)} & \scriptsize{(0.342, 0.396)} & \scriptsize{(1.14, 1.57)} & \scriptsize{(0.928, 0.934)} & \scriptsize{(3.98, 7.68)} \\ [.5ex]
\hline \\ [-2.3ex]
\multirow{2}{*}{\instset{Direct}} & 0.937 & \textbf{0.366} & 0.145 & \textbf{0.335} & \textbf{1.34$\mathbf{\times10^{-4}}$} & 0.934 & 3.98$\times10^{-3}$ \\ 
 & \scriptsize{(0.916, 0.943)} & \scriptsize{(0.329, 0.416)} & \scriptsize{(0.144, 0.146)} & \scriptsize{(0.295, 0.383)} & \scriptsize{(1.10, 1.57)} & \scriptsize{(0.931, 0.936)} & \scriptsize{(3.97, 6.16)} \\ [.5ex]
\hline \\ [-2.3ex]
\multirow{2}{*}{\instset{Both}} & 0.922 & \textbf{0.422} & 0.145 & \textbf{0.388} & \textbf{1.37$\mathbf{\times10^{-4}}$} & 0.932 & 7.05$\times10^{-3}$ \\ 
 & \scriptsize{(0.857, 0.939)} & \scriptsize{(0.385, 0.495)} & \scriptsize{(0.140, 0.146)} & \scriptsize{(0.367, 0.403)} & \scriptsize{(1.09, 1.51)} & \scriptsize{(0.929, 0.935)} & \scriptsize{(4.00, 7.96)} \\ [.5ex]
\hline \\ [-2.3ex]
\multirow{2}{*}{\instset{Seq}} & 0.932 & 0.509 & 0.143 & \textbf{0.397} & 1.68$\times10^{-4}$ & 0.928 & 5.31$\times10^{-3}$ \\ 
 & \scriptsize{(0.919, 0.938)} & \scriptsize{(0.460, 0.573)} & \scriptsize{(0.139, 0.144)} & \scriptsize{(0.365, 0.405)} & \scriptsize{(1.57, 2.02)} & \scriptsize{(0.925, 0.929)} & \scriptsize{(3.98, 7.57)} \\ [.5ex]
\hline \\ [-2.3ex]
\instset{Seq} & 0.929 & 0.559 & 0.144 & \textbf{0.370} & 1.61$\times10^{-4}$ & 0.931 & 5.06$\times10^{-3}$ \\ 
\instset{Direct} & \scriptsize{(0.918, 0.939)} & \scriptsize{(0.522, 0.612)} & \scriptsize{(0.141, 0.146)} & \scriptsize{(0.309, 0.398)} & \scriptsize{(1.49, 1.81)} & \scriptsize{(0.928, 0.934)} & \scriptsize{(3.98, 7.70)} \\ [.5ex]
\hline \\ [-2.3ex]
\instset{Direct} & 0.932 & 0.542 & 0.143 & \textbf{0.359} & 1.53$\times10^{-4}$ & 0.930 & 5.24$\times10^{-3}$ \\ 
\instset{Seq} & \scriptsize{(0.919, 0.941)} & \scriptsize{(0.500, 0.562)} & \scriptsize{(0.141, 0.145)} & \scriptsize{(0.300, 0.399)} & \scriptsize{(1.36, 1.77)} & \scriptsize{(0.928, 0.933)} & \scriptsize{(3.98, 7.63)} \\ [.5ex]
\hline \\ [-2.3ex]
\instset{Seq} & 0.926 & 0.517 & \textbf{0.079} & \textbf{0.374} & 1.64$\times10^{-4}$ & 0.928 & 7.74$\times10^{-3}$ \\ 
\instset{Both} & \scriptsize{(0.914, 0.934)} & \scriptsize{(0.482, 0.545)} & \scriptsize{(0.078, 0.125)} & \scriptsize{(0.300, 0.398)} & \scriptsize{(1.51, 1.94)} & \scriptsize{(0.923, 0.930)} & \scriptsize{(7.20, 7.91)} \\ [.5ex]
\hline\hline
\end{tabular*} \\ [0.5ex]
\begin{flushleft}
Task success results of the \instset{Label}-series instruction set architectures.  Each entry shows the median normalized task success in the respective environment, with $95\%$ confidence intervals in parentheses. Bold entries denote significant ($p < 0.05$, Wilcoxon rank-sum test) deviations.
\end{flushleft}
\end{table}

\begin{table}[!ht]
\begin{flushleft}
\caption{\label{tbl:is:splitio:fit}bf \instset{Split-IO} Architecture Fitness}
\end{flushleft}
\begin{tabular*}{\hsize}{@{\extracolsep{\fill}} l l l l l l l l}
\\ [-2ex]
\hline\hline \\ [-2.3ex]
& Logic-9 & Logic-77 & Match-12 & Fib.-32 & Sort-10 & Limited-9 & Navigation \\ [0.1ex]
\hline \\ [-2.3ex]
\instset{Seq} & 22.56 & 43.57 & 0.207 & 6.106 & -0.32 & 5.812 & 2.641 \\ 
\instset{Direct} & \scriptsize{(22.32, 22.72)} & \scriptsize{(39.73, 46.60)} &  \scriptsize{(0.182, 0.239)} & \scriptsize{(5.326, 6.549)} & \scriptsize{(-0.39, -0.24)} & \scriptsize{(5.390, 6.169)} & \scriptsize{(1.198, 3.341)} \\ [.5ex]
\hline \\ [-2.3ex]
\multirow{2}{*}{\instset{SplitIO}} & \textbf{23.07} & \textbf{53.94} & \textbf{0.337} & \textbf{8.096} & \textbf{-1.03} & 5.343 & 1.091 \\ 
 & \scriptsize{(22.87, 23.22)} & \scriptsize{(50.34, 56.72)} &  \scriptsize{(0.314, 0.360)} & \scriptsize{(7.983, 8.207)} & \scriptsize{(-1.03, -1.02)} & \scriptsize{(5.221, 5.520)} & \scriptsize{(1.062, 2.920)} \\ [.5ex]
\hline\hline
\end{tabular*} \\ [0.5ex]
\begin{flushleft}
Fitness results of the \instset{Label-Seq-Direct} and \instset{Split-IO} instruction set architectures. Each entry shows the median log$_2$ population mean fitness in the respective environment, with $95\%$ confidence intervals in parentheses. Bold entries indicate significant ($p < 0.05$, Wilcoxon rank-sum test) deviations after sequential Bonferroni correction.
\end{flushleft}
\end{table}

\begin{table}[!ht]
\begin{flushleft}
\caption{\label{tbl:is:splitio:ts}\bf \instset{Split-IO} Architecture Task Success}
\end{flushleft}
\begin{tabular*}{\hsize}{@{\extracolsep{\fill}} l l l l l l l l}
\\ [-2ex]
\hline\hline \\ [-2.3ex]
& Logic-9 & Logic-77 & Match-12 & Fib.-32 & Sort-10 & Limited-9 & Navigation \\ [0.1ex]
\hline \\ [-2.3ex]
\instset{Seq} & 0.930 & 0.559 & 0.145 & 0.384 & 1.62$\times10^{-4}$ & 0.926 & 6.63$\times10^{-3}$ \\ 
\instset{Direct} & \scriptsize{(0.920, 0.935)} & \scriptsize{(0.521, 0.593)} & \scriptsize{(0.142, 0.146)} & \scriptsize{(0.318, 0.397)} & \scriptsize{(1.52, 1.74)} & \scriptsize{(0.922, 0.928)} & \scriptsize{(3.99, 7.94)} \\ [.5ex]
\hline \\ [-2.3ex]
\multirow{2}{*}{\instset{SplitIO}} & \textbf{0.940} & \textbf{0.678} & \textbf{0.148} & \textbf{0.449} & \textbf{0.0} & \textbf{0.931} & 3.99$\times10^{-3}$ \\ 
 & \scriptsize{(0.936, 0.942)} & \scriptsize{(0.651, 0.707)} & \scriptsize{(0.148, 0.149)} & \scriptsize{(0.447, 0.461)} & \scriptsize{(0.0, 0.0)} & \scriptsize{(0.927, 0.933)} & \scriptsize{(3.97, 7.41)} \\ [.5ex]
\hline\hline
\end{tabular*} \\ [0.5ex]
\begin{flushleft}
Task success results of the \instset{Label-Seq-Direct} and \instset{Split-IO} instruction set architectures.  Each entry shows the median normalized task success in the respective environment, with $95\%$ confidence intervals in parentheses. Bold entries denote significant ($p < 0.05$, Wilcoxon rank-sum test) deviations.
\end{flushleft}
\end{table}

\begin{table}[!ht]
\begin{flushleft}
\caption{\label{tbl:is:search:fit}\bf \instset{Search} Series Architectures Fitness}
\end{flushleft}
\begin{tabular*}{\hsize}{@{\extracolsep{\fill}} l l l l l l l l}
\\ [-2ex]
\hline\hline \\ [-2.3ex]
& Logic-9 & Logic-77 & Match-12 & Fib.-32 & Sort-10 & Limited-9 & Navigation \\ [0.1ex]
\hline \\ [-2.3ex]
\multirow{2}{*}{\instset{SplitIO}} & 23.19 & 54.58 & 0.307 & 8.139 & -1.03 & 5.477 & 2.909 \\ 
 & \scriptsize{(23.05, 23.25)} & \scriptsize{(52.46, 58.54)} &  \scriptsize{(0.261, 0.335)} & \scriptsize{(8.027, 8.318)} & \scriptsize{(-1.04, -1.02)} & \scriptsize{(5.014, 5.912)} & \scriptsize{(1.121, 3.382)} \\ [.5ex]
\hline \\ [-2.3ex]
\multirow{2}{*}{\instset{Search}} & 23.02 & \textbf{48.75} & 0.313 & 8.188 & -1.02 & 5.393 & 3.150 \\ 
 & \scriptsize{(22.87, 23.17)} & \scriptsize{(46.33, 52.21)} &  \scriptsize{(0.265, 0.335)} & \scriptsize{(8.042, 8.273)} & \scriptsize{(-1.03, -0.98)} & \scriptsize{(5.177, 5.745)} & \scriptsize{(1.708, 3.431)} \\ [.5ex]
\hline \\ [-2.3ex]
\multirow{2}{*}{\instset{Goto}} & 23.13 & \textbf{50.42} & 0.311 & 7.946 & -1.04 & 5.598 & 2.584 \\ 
 & \scriptsize{(22.90, 23.21)} & \scriptsize{(48.27, 53.34)} &  \scriptsize{(0.232, 0.337)} & \scriptsize{(7.853, 8.080)} & \scriptsize{(-1.05, -1.02)} & \scriptsize{(5.272, 5.850)} & \scriptsize{(1.084, 3.219)} \\ [.5ex]
\hline \\ [-2.3ex]
\multirow{2}{*}{\instset{GotoIf}} & \textbf{22.92} & \textbf{48.49} & 0.283 & 7.937 & -1.04 & 5.840 & 2.283 \\ 
 & \scriptsize{(22.61, 23.06)} & \scriptsize{(44.61, 52.01)} &  \scriptsize{(0.223, 0.336)} & \scriptsize{(7.844, 8.070)} & \scriptsize{(-1.05, -1.01)} & \scriptsize{(5.624, 6.059)} & \scriptsize{(1.322, 3.028)} \\ [.5ex]
\hline\hline
\end{tabular*} \\ [0.5ex]
\begin{flushleft}
Fitness results of the \instset{Search}-series instruction set architectures. Each entry shows the median log$_2$ population mean fitness in the respective environment, with $95\%$ confidence intervals in parentheses. Bold entries indicate significant ($p < 0.05$, Wilcoxon rank-sum test) deviations after sequential Bonferroni correction.
\end{flushleft}
\end{table}

\begin{table}[!ht]
\begin{flushleft}
\caption{\label{tbl:is:search:ts}\bf \instset{Search} Series Architectures Task Success}
\end{flushleft}
\begin{tabular*}{\hsize}{@{\extracolsep{\fill}} l l l l l l l l}
\\ [-2ex]
\hline\hline \\ [-2.3ex]
& Logic-9 & Logic-77 & Match-12 & Fib.-32 & Sort-10 & Limited-9 & Navigation \\ [0.1ex]
\hline \\ [-2.3ex]
\multirow{2}{*}{\instset{SplitIO}} & 0.937 & 0.694 & 0.149 & 0.449 & 0.0 & 0.927 & 7.28$\times10^{-3}$ \\ 
 & \scriptsize{(0.934, 0.942)} & \scriptsize{(0.658, 0.719)} & \scriptsize{(0.148, 0.149)} & \scriptsize{(0.446, 0.463)} & \scriptsize{(0.0, 0.0)} & \scriptsize{(0.926, 0.930)} & \scriptsize{(3.99, 8.03)} \\ [.5ex]
\hline \\ [-2.3ex]
\multirow{2}{*}{\instset{Search}} & \textbf{0.937} & 0.626 & 0.149 & 0.448 & \textbf{0.0} & 0.929 & 7.67$\times10^{-3}$ \\ 
 & \scriptsize{(0.932, 0.941)} & \scriptsize{(0.584, 0.652)} & \scriptsize{(0.148, 0.150)} & \scriptsize{(0.445, 0.455)} & \scriptsize{(0.0, 0.0)} & \scriptsize{(0.927, 0.932)} & \scriptsize{(4.60, 8.01)} \\ [.5ex]
\hline \\ [-2.3ex]
\multirow{2}{*}{\instset{Goto}} & 0.940 & \textbf{0.644} & 0.148 & 0.447 & \textbf{0.0} & 0.930 & 6.43$\times10^{-3}$ \\ 
 & \scriptsize{(0.935, 0.943)} & \scriptsize{(0.608, 0.676)} & \scriptsize{(0.147, 0.149)} & \scriptsize{(0.444, 0.452)} & \scriptsize{(0.0, 0.0)} & \scriptsize{(0.928, 0.933)} & \scriptsize{(3.99, 7.61)} \\ [.5ex]
\hline \\ [-2.3ex]
\multirow{2}{*}{\instset{GotoIf}} & \textbf{0.933} & \textbf{0.601} & 0.149 & 0.447 & \textbf{0.0} & 0.929 & 5.78$\times10^{-3}$ \\ 
 & \scriptsize{(0.928, 0.939)} & \scriptsize{(0.569, 0.650)} & \scriptsize{(0.148, 0.150)} & \scriptsize{(0.445, 0.448)} & \scriptsize{(0.0, 0.0)} & \scriptsize{(0.925, 0.933)} & \scriptsize{(4.15, 7.57)} \\ [.5ex]
\hline\hline
\end{tabular*} \\ [0.5ex]
\begin{flushleft}
Task success results of the \instset{Search}-series instruction set architectures.  Each entry shows the median normalized task success in the respective environment, with $95\%$ confidence intervals in parentheses. Bold entries denote significant ($p < 0.05$, Wilcoxon rank-sum test) deviations.
\end{flushleft}
\end{table}

\begin{table}[!ht]
\begin{flushleft}
\caption{\label{tbl:is:flow:fit}\bf \instset{Flow} Series Architectures Fitness}
\end{flushleft}
\begin{tabular*}{\hsize}{@{\extracolsep{\fill}} l l l l l l l l}
\\ [-2ex]
\hline\hline \\ [-2.3ex]
& Logic-9 & Logic-77 & Match-12 & Fib.-32 & Sort-10 & Limited-9 & Navigation \\ [0.1ex]
\hline \\ [-2.3ex]
\multirow{2}{*}{\instset{Search}} & 23.14 & 48.59 & 0.313 & 8.061 & -1.02 & 5.571 & 3.022 \\ 
 & \scriptsize{(23.01, 23.23)} & \scriptsize{(46.46, 51.41)} &  \scriptsize{(0.243, 0.346)} & \scriptsize{(7.990, 8.176)} & \scriptsize{(-1.04, -1.01)} & \scriptsize{(5.334, 5.867)} & \scriptsize{(2.151, 3.387)} \\ [.5ex]
\hline \\ [-2.3ex]
\multirow{2}{*}{\instset{MvH}} & \textbf{22.78} & \textbf{42.25} & 0.277 & 8.124 & \textbf{-0.96} & 5.474 & \textbf{3.875} \\ 
 & \scriptsize{(22.38, 23.05)} & \scriptsize{(39.76, 46.35)} &  \scriptsize{(0.216, 0.347)} & \scriptsize{(7.990, 8.270)} & \scriptsize{(-1.01, -0.69)} & \scriptsize{(5.181, 5.771)} & \scriptsize{(3.729, 4.052)} \\ [.5ex]
\hline \\ [-2.3ex]
\multirow{2}{*}{\instset{If0}} & \textbf{22.80} & 47.99 & 0.296 & 7.995 & -1.04 & 5.553 & 3.229 \\ 
 & \scriptsize{(22.62, 23.07)} & \scriptsize{(45.02, 51.50)} &  \scriptsize{(0.258, 0.323)} & \scriptsize{(7.857, 8.099)} & \scriptsize{(-1.05, -1.02)} & \scriptsize{(5.326, 5.855)} & \scriptsize{(2.919, 3.549)} \\ [.5ex]
\hline \\ [-2.3ex]
\multirow{2}{*}{\instset{IfX}} & \textbf{22.64} & 46.40 & \textbf{0.199} & 8.037 & \textbf{-1.06} & 5.804 & \textbf{4.028} \\ 
 & \scriptsize{(22.33, 22.86)} & \scriptsize{(44.51, 48.76)} &  \scriptsize{(0.168, 0.264)} & \scriptsize{(7.964, 8.198)} & \scriptsize{(-1.07, -1.03)} & \scriptsize{(5.460, 6.243)} & \scriptsize{(3.861, 4.312)} \\ [.5ex]
\hline \\ [-2.3ex]
\multirow{2}{*}{\instset{If0-IfX}} & \textbf{22.82} & 46.00 & \textbf{0.270} & 8.011 & \textbf{-1.08} & 5.595 & \textbf{4.331} \\ 
 & \scriptsize{(22.65, 23.01)} & \scriptsize{(42.64, 49.03)} &  \scriptsize{(0.201, 0.308)} & \scriptsize{(7.952, 8.078)} & \scriptsize{(-1.09, -1.07)} & \scriptsize{(5.292, 6.071)} & \scriptsize{(4.113, 4.615)} \\ [.5ex]
\hline \\ [-2.3ex]
\instset{IfX} & \textbf{22.95} & \textbf{41.29} & 0.311 & 8.063 & -1.00 & 5.751 & \textbf{4.475} \\ 
\instset{MvH} & \scriptsize{(22.74, 23.11)} & \scriptsize{(38.48, 44.64)} &  \scriptsize{(0.244, 0.346)} & \scriptsize{(7.980, 8.193)} & \scriptsize{(-1.03, -0.92)} & \scriptsize{(5.522, 6.061)} & \scriptsize{(4.244, 4.983)} \\ [.5ex]
\hline \\ [-2.3ex]
\instset{If0-IfX} & \textbf{21.94} & \textbf{41.76} & \textbf{0.213} & 7.995 & -1.01 & 6.077 & \textbf{5.036} \\ 
\instset{MvH} & \scriptsize{(21.55, 22.37)} & \scriptsize{(39.12, 43.84)} &  \scriptsize{(0.189, 0.283)} & \scriptsize{(7.849, 8.066)} & \scriptsize{(-1.05, -0.91)} & \scriptsize{(6.723, 6.625)} & \scriptsize{(4.576, 6.457)} \\ [.5ex]
\hline\hline
\end{tabular*} \\ [0.5ex]
\begin{flushleft}
Fitness results of the \instset{Flow}-series instruction set architectures. Each entry shows the median log$_2$ population mean fitness in the respective environment, with $95\%$ confidence intervals in parentheses. Bold entries indicate significant ($p < 0.05$, Wilcoxon rank-sum test) deviations after sequential Bonferroni correction.
\end{flushleft}
\end{table}

\begin{table}[!ht]
\begin{flushleft}
\caption{\label{tbl:is:flow:ts}\bf \instset{Flow} Series Architectures Task Success}
\end{flushleft}
\begin{tabular*}{\hsize}{@{\extracolsep{\fill}} l l l l l l l l}
\\ [-2ex]
\hline\hline \\ [-2.3ex]
& Logic-9 & Logic-77 & Match-12 & Fib.-32 & Sort-10 & Limited-9 & Navigation \\ [0.1ex]
\hline \\ [-2.3ex]
\multirow{2}{*}{\instset{Search}} & 0.943 & 0.623 & 0.148 & 0.448 & 0.0 & 0.928 & 7.55$\times10^{-3}$ \\ 
 & \scriptsize{(0.939, 0.946)} & \scriptsize{(0.584, 0.648)} & \scriptsize{(0.147, 0.149)} & \scriptsize{(0.444, 0.452)} & \scriptsize{(0.0, 0.0)} & \scriptsize{(0.926, 0.932)} & \scriptsize{(5.51, 8.04)} \\ [.5ex]
\hline \\ [-2.3ex]
\multirow{2}{*}{\instset{MvH}} & 0.938 & 0.563 & 0.150 & 0.465 & \textbf{0.0} & \textbf{0.945} & \textbf{8.63$\mathbf{\times10^{-3}}$} \\ 
 & \scriptsize{(0.930, 0.946)} & \scriptsize{(0.532, 0.593)} & \scriptsize{(0.149, 0.150)} & \scriptsize{(0.452, 0.476)} & \scriptsize{(0.0, 0.0)} & \scriptsize{(0.941, 0.948)} & \scriptsize{(8.40, 8.93)} \\ [.5ex]
\hline \\ [-2.3ex]
\multirow{2}{*}{\instset{If0}} & \textbf{0.932} & 0.611 & 0.148 & 0.447 & 0.0 & 0.932 & 7.78$\times10^{-3}$ \\ 
 & \scriptsize{(0.924, 0.937)} & \scriptsize{(0.581, 0.647)} & \scriptsize{(0.147, 0.149)} & \scriptsize{(0.446, 0.451)} & \scriptsize{(0.0, 0.0)} & \scriptsize{(0.930, 0.935)} & \scriptsize{(6.52, 8.12)} \\ [.5ex]
\hline \\ [-2.3ex]
\multirow{2}{*}{\instset{IfX}} & \textbf{0.929} & 0.607 & 0.147 & 0.447 & 0.0 & 0.931 & \textbf{8.88$\mathbf{\times10^{-3}}$} \\ 
 & \scriptsize{(0.916, 0.935)} & \scriptsize{(0.577, 0.622)} & \scriptsize{(0.146, 0.148)} & \scriptsize{(0.445, 0.453)} & \scriptsize{(0.0, 0.0)} & \scriptsize{(0.929, 0.933)} & \scriptsize{(8.43, 9.67)} \\ [.5ex]
\hline \\ [-2.3ex]
\multirow{2}{*}{\instset{If0-IfX}} & 0.937 & 0.594 & 0.148 & 0.451 & 0.0 & 0.931 & \textbf{9.47$\mathbf{\times10^{-3}}$} \\ 
 & \scriptsize{(0.932, 0.941)} & \scriptsize{(0.550, 0.626)} & \scriptsize{(0.147, 0.149)} & \scriptsize{(0.446, 0.459)} & \scriptsize{(0.0, 0.0)} & \scriptsize{(0.928, 0.935)} & \scriptsize{(8.75, 10.12)} \\ [.5ex]
\hline \\ [-2.3ex]
\instset{IfX} & 0.945 & \textbf{0.549} & \textbf{0.150} & 0.459 & \textbf{0.0} & \textbf{0.941} & \textbf{9.65$\mathbf{\times10^{-3}}$} \\ 
\instset{MvH} & \scriptsize{(0.940, 0.951)} & \scriptsize{(0.508, 0.576)} & \scriptsize{(0.149, 0.151)} & \scriptsize{(0.451, 0.467)} & \scriptsize{(0.0, 0.0)} & \scriptsize{(0.937, 0.947)} & \scriptsize{(9.10, 10.64)} \\ [.5ex]
\hline \\ [-2.3ex]
\instset{If0-IfX} & \textbf{0.926} & \textbf{0.544} & 0.150 & 0.451 & \textbf{0.0} & \textbf{0.940} & \textbf{11.25$\mathbf{\times10^{-3}}$} \\ 
\instset{MvH} & \scriptsize{(0.901, 0.935)} & \scriptsize{(0.516, 0.577)} & \scriptsize{(0.148, 0.150)} & \scriptsize{(0.447, 0.459)} & \scriptsize{(0.0, 0.0)} & \scriptsize{(0.935, 0.945)} & \scriptsize{(10.23, 13.40)} \\ [.5ex]
\hline\hline
\end{tabular*} \\ [0.5ex]
\begin{flushleft}
Task success results of the \instset{Flow}-series instruction set architectures.  Each entry shows the median normalized task success in the respective environment, with $95\%$ confidence intervals in parentheses. Bold entries denote significant ($p < 0.05$, Wilcoxon rank-sum test) deviations.
\end{flushleft}
\end{table}

\begin{table}[!ht]
\begin{flushleft}
\caption{\label{tbl:is:final:fit}\bf \instset{Heads} and \instset{Heads-EX} Architectures Fitness}
\end{flushleft}
\begin{tabular*}{\hsize}{@{\extracolsep{\fill}} l l l l l l l l}
\\ [-2ex]
\hline\hline \\ [-2.3ex]
& Logic-9 & Logic-77 & Match-12 & Fib.-32 & Sort-10 & Limited-9 & Navigation \\ [0.1ex]
\hline \\ [-2.3ex]
\multirow{2}{*}{\instset{Heads}} & 19.44 & 13.50 & 0.194 & 3.453 & -0.47 & 4.328 & 1.656 \\ 
 & \scriptsize{(17.74, 19.79)} & \scriptsize{(11.67, 15.30)} &  \scriptsize{(0.168, 0.248)} & \scriptsize{(3.216, 3.858)} & \scriptsize{(-0.61, -0.33)} & \scriptsize{(4.157, 4.445)} & \scriptsize{(1.108, 3.606)} \\ [.5ex]
\hline \\ [-2.3ex]
\instset{IfX} & \textbf{22.95} & \textbf{41.292} & \textbf{0.311} & \textbf{8.063} & \textbf{-1.00} & \textbf{5.751} & \textbf{4.475} \\ 
\instset{MvH} & \scriptsize{(22.74, 23.11)} & \scriptsize{(38.50, 44.56)} &  \scriptsize{(0.245, 0.347)} & \scriptsize{(7.980, 8.189)} & \scriptsize{(-1.03, -0.92)} & \scriptsize{(5.517, 6.049)} & \scriptsize{(4.244, 4.953)} \\ [.5ex]
\hline\hline
\end{tabular*} \\ [0.5ex]
\begin{flushleft}
Fitness results for the base \instset{Heads} and the \instset{Heads-EX} instruction set architectures. The \instset{Heads-EX} architecture includes features from all six tested feature groups, including fully associative arguments, six registers, direct-matched labels, split-I/O, directional search instructions, the \instr{ifx} instruction, and conditional \instr{mov-head} instructions.  Each entry shows the median log$_2$ population mean fitness in the respective environment, with $95\%$ confidence intervals in parentheses. Bold entries indicate significant ($p < 0.05$, Wilcoxon rank-sum test) deviations.
\end{flushleft}
\end{table}

\begin{table}[!ht]
\begin{flushleft}
\caption{\label{tbl:is:final:ts}\bf \instset{Heads} and \instset{Heads-EX} Architectures Task Success}
\end{flushleft}
\begin{tabular*}{\hsize}{@{\extracolsep{\fill}} l l l l l l l l}
\\ [-2ex]
\hline\hline \\ [-2.3ex]
& Logic-9 & Logic-77 & Match-12 & Fib.-32 & Sort-10 & Limited-9 & Navigation \\ [0.1ex]
\hline \\ [-2.3ex]
\multirow{2}{*}{\instset{Heads}} & 0.834 & 0.185 & 0.146 & 0.202 & 1.42$\times10^{-4}$ & 0.908 & 4.72$\times10^{-3}$ \\ 
 & \scriptsize{(0.752, 0.844)} & \scriptsize{(0.162, 0.211)} & \scriptsize{(0.145, 0.147)} & \scriptsize{(0.177, 0.228)} & \scriptsize{(1.08, 1.66)} & \scriptsize{(0.897, 0.914)} & \scriptsize{(3.99, 8.23)} \\ [.5ex]
\hline \\ [-2.3ex]
\instset{IfX} & \textbf{0.945} & \textbf{0.549} & \textbf{0.150} & \textbf{0.459} & \textbf{0.0} & \textbf{0.941} & \textbf{9.65$\mathbf{\times10^{-3}}$} \\ 
\instset{MvH} & \scriptsize{(0.940, 0.951)} & \scriptsize{(0.507, 0.577)} & \scriptsize{(0.149, 0.151)} & \scriptsize{(0.451, 0.467)} & \scriptsize{(0.0, 0.0)} & \scriptsize{(0.937, 0.947)} & \scriptsize{(9.12, 10.62)} \\ [.5ex]
\hline\hline
\end{tabular*} \\ [0.5ex]
\begin{flushleft}
Task success results for the base \instset{Heads} and the \instset{Heads-EX} instruction set architectures. The \instset{Heads-EX} architecture includes features from all six tested feature groups, including fully associative arguments, six registers, direct-matched labels, split-I/O, directional search instructions, the \instr{ifx} instruction, and conditional \instr{mov-head} instructions. Each entry shows the median normalized task success in the respective environment, with $95\%$ confidence intervals in parentheses. Bold entries denote significant ($p < 0.05$, Wilcoxon rank-sum test) deviations.
\end{flushleft}
\end{table}

\end{document}